\documentclass{article}

    \PassOptionsToPackage{numbers, compress}{natbib}

    \usepackage[preprint]{neurips_2024}

\usepackage[utf8]{inputenc} 
\usepackage[T1]{fontenc}    
\usepackage{hyperref}       
\usepackage{url}            
\usepackage{booktabs}       
\usepackage{amsfonts}       
\usepackage{nicefrac}       
\usepackage{microtype}      
\usepackage{xcolor}         

\usepackage{amsmath,amssymb} 
\usepackage{color}
\usepackage{wrapfig}
\usepackage{multirow}

\usepackage{graphicx}
\usepackage{pythonhighlight}

\usepackage{booktabs} 
\usepackage{colortbl}

\usepackage{svg}

\usepackage{enumitem}
\setitemize{itemsep=10pt,topsep=0pt,parsep=0pt,partopsep=0pt}

\DeclareMathOperator*{\argmax}{arg\,max}

\definecolor{red}{RGB}{255,0,0}

\title{MiniCPM-V: A GPT-4V Level MLLM on Your Phone}

\definecolor{mypink}{RGB}{239,43,159}

\author{
Yuan Yao$\thanks{\hspace{0.1em}Core team, $^{\dag}$ Project lead, $^\ddag$ Corresponding author } \hspace{0.45em} ^{\dagger}$ \ \ \
Tianyu Yu$^{*}$ \ \
Ao Zhang$^{*}$ \ \ 
Chongyi Wang$^{*}$ \ \ 
Junbo Cui$^{*}$   \ \ 
Hongji Zhu$^{*}$ \ \ \\
\textbf{Tianchi Cai$^{*}$ \ \
Haoyu Li \ \
Weilin Zhao \ \
Zhihui He \ \
Qianyu Chen \ \ 
Huarong Zhou \ \ 
\ \
}\\
\textbf{
Zhensheng Zou \ \  
Haoye Zhang \ \
Shengding Hu\ \
Zhi Zheng \ \
Jie Zhou \ \ 
Jie Cai \ \ 
} \\
\textbf{Xu Han \ \
Guoyang Zeng \ \
Dahai Li \ \
Zhiyuan Liu \ \ 
Maosong Sun$^{\ddag}$} \ \
\\[0.5em]
MiniCPM-V Team, OpenBMB
\ \ \ \\
{\texttt{yaoyuanthu@gmail.com}} \\
\\
\large{\color{mypink} \textbf{ \url{https://github.com/OpenBMB/MiniCPM-V}}}
}

\begin{document}

\maketitle

\begin{figure*}[h]
\centering
\includegraphics[width=1.0\linewidth]{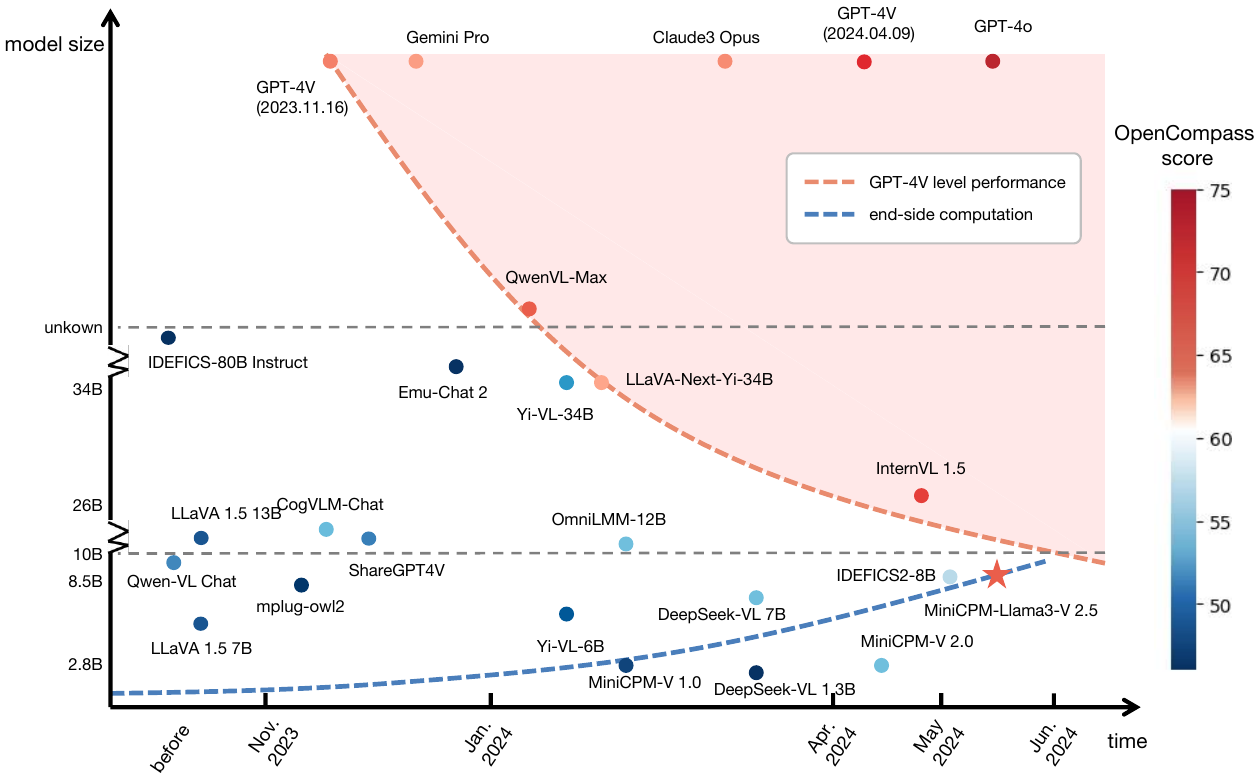}
\caption{Moore's Law for MLLM? Trends of MLLM development in terms of time (x-axis), model size (y-axis), and performance (color). The \textbf{red line} shows the decreasing model sizes for achieving GPT-4V level performance, while the \textbf{blue line} represents the growing end-side computation capacity. This jointly shows that GPT-4V level MLLMs deployed on end devices are becoming increasingly possible, unlocking a wider spectrum of real-world AI applications in the near future.}
\label{fig:teaser}
\end{figure*}

\begin{abstract}
The recent surge of Multimodal Large Language Models (MLLMs) has fundamentally reshaped the landscape of AI research and industry, shedding light on a promising path toward the next AI milestone. However, significant challenges remain preventing MLLMs from being practical in real-world applications. The most notable challenge comes from the huge cost of running an MLLM with a massive number of parameters and extensive computation. As a result, most MLLMs need to be deployed on high-performing cloud servers, which greatly limits their application scopes such as mobile, offline, energy-sensitive, and privacy-protective scenarios. In this work, we present MiniCPM-V, a series of efficient MLLMs deployable on end-side devices. By integrating the latest MLLM techniques in architecture, pretraining and alignment, the latest MiniCPM-Llama3-V 2.5 has several notable features: (1) Strong performance, outperforming GPT-4V-1106, Gemini Pro and Claude 3 on OpenCompass, a comprehensive evaluation over 11 popular benchmarks, (2) strong OCR capability and 1.8M pixel high-resolution image perception at any aspect ratio, (3) trustworthy behavior with low hallucination rates, (4) multilingual support for 30+ languages, and (5) efficient deployment on mobile phones. More importantly, MiniCPM-V can be viewed as a representative example of a promising trend (Fig.~\ref{fig:teaser}): The model sizes for achieving usable (e.g., GPT-4V) level performance are rapidly decreasing, along with the fast growth of end-side computation capacity. This jointly shows that GPT-4V level MLLMs deployed on end devices are becoming increasingly possible, unlocking a wider spectrum of real-world AI applications in the near future.

\end{abstract}

\section{Introduction}
The rapid development of Multimodal Large Language Models (MLLMs)~\citep{reid2024gemini, achiam2023gpt4, lu2024deepseekvl, liu2023llava, wang2023cogvlm, chen2024far, qwenvl, hu2023viscpm, idefics2,mckinzie2024mm1,beyer2024paligemma} have brought an impressive surge in multimodal capabilities in understanding, reasoning and interaction. This has not only fundamentally reshaped the landscape of AI research and industry, but also shed light on a promising path towards the next AI milestone. However, current MLLMs are still far from being practical in real-world applications. One of the most predominant challenges is that current MLLMs typically entail a massive number of parameters and impose heavy computational burdens. As a result, most MLLMs can only be deployed on high-performing cloud servers, leading to significant energy consumption and carbon emissions. This limitation significantly constrains the potential application scopes such as on mobile devices, energy-sensitive scenarios, offline scenarios without stable network connections, and privacy/security protective scenarios for both personal and industrial users.

In light of these limitations, there is a growing interest in exploring more efficient lightweight MLLMs~\citep{abdin2024phi3, lu2024deepseekvl, reid2024gemini, beyer2024paligemma} that can run on end-side devices. End-side scenarios encompass a broader scope of equipment, including mobile phones, personal computers, vehicles and robotics, etc., which are ubiquitous in users' daily lives and experiencing rapid advancements in computation capacities. End-side MLLMs provide a promising solution towards more practical applications due to their broader usage scope, better computation efficiency, more robust offline behaviors, and better privacy/security protection.

However, developing capable end-side MLLMs is challenging due to significantly constrained parameter and inference computation budgets. As a result, more careful architecture designs and training recipes are required to fully unleash the potential of end-side MLLMs. In this work, we present MiniCPM-V, a series of efficient MLLMs deployable on end-side devices. The philosophy of MiniCPM-V is to achieve a good balance between performance and efficiency, a more important objective in real-world applications. To date in 2024, we have unveiled three models: (1) In February, we launched MiniCPM-V 1.0 2B, one of the first MLLMs designed for mobile phones. (2) In April, MiniCPM-V 2.0 2B was introduced, outperforming strong larger MLLMs such as Qwen-VL 9B~\cite{qwenvl}, CogVLM 17B~\cite{wang2023cogvlm}, and Yi-VL 34B~\citep{ai2024yi}. This iteration also introduces support for high-resolution image input and exhibits promising OCR capabilities. (3) Most recently in May, we released MiniCPM-Llama3-V 2.5 8B, which outperforms strong GPT-4V-1106, Gemini Pro and Claude 3 on OpenCompass evaluation. Noteworthy features of this model include strong OCR capability, high-resolution image perception, trustworthy behavior, multilingual support, and efficient end-side deployment optimization.

More importantly, MiniCPM-V can be viewed as a representative example of a promising trend. Fig.~\ref{fig:teaser} summarizes the recent development of MLLMs~\citep{abdin2024phi3, lu2024deepseekvl, li2024minigemini} in terms of performance, parameters and release time. 
We observe an interesting trend akin to Moore's Law~\citep{moorelaw} indicated by the red line: the sizes of models reaching the GPT-4V level performance are rapidly decreasing over time. This phenomenon could perhaps be called the \textit{Moore's Law of MLLMs}. Simultaneously, the computational capacity of end-side devices such as phones and personal computers is steadily increasing (qualitatively depicted by the blue line).
The convergence of these two trends indicates usable (e.g., GPT-4V level) MLLMs deployable on end-side devices are soon within reach, opening up broader possibilities and benefiting more application scenarios in the near future.
From a historical perspective of human technology development, this trend can also be viewed as human pursuit of miniaturization of state-of-the-art technologies, which have been repeatedly witnessed in other science and technology fields. For example, in aerospace, the latest SpaceX Raptor 2 rocket engine can achieve a strong thrust of 2,256 kN with a mass of 1.6 tons, whereas 20 years ago, the RD-0750 rocket engine could only achieve a thrust of 1,413 kN with a mass exceeding 4 tons~\citep{thrust}.

\paragraph{MiniCPM-V Series Techniques.}

In this paper, we will take the latest MiniCPM-Llama3-V 2.5 as an example, and systematically introduce the notable features of MiniCPM-V series and the key techniques behind them:
\begin{itemize}[leftmargin=7.5mm]
\setlength{\itemsep}{2pt}
\item  {\textbf{Leading Performance}}.
MiniCPM-Llama3-V 2.5 achieves better performance than GPT-4V-1106, Gemini Pro and Claude 3 on OpenCompass collection, a comprehensive evaluation over 11 popular benchmarks. This is jointly contributed by its careful design in architecture, data and training recipes, which we will detail in the following.  

\item  {\textbf{Strong OCR Capability.}} MiniCPM-Llama3-V 2.5 outperforms GPT-4V, Gemini Pro and Qwen-VL-Max on OCRBench. It also supports high-utility functions such as table-to-markdown conversion and full OCR content transcribtion. These are largely attributed to the 1.8M pixel high-resolution (e.g., 1344 $\times$ 1344) image perception technique across any aspect ratios~\cite{xu2024llavauhd}.
\item  {\textbf{Trustworthy Behavior.}} Based on the RLAIF-V~\citep{yu2022rlaifv} and RLHF-V~\cite{yu2024rlhf} techniques that align MLLM behaviors from AI/human feedback, MiniCPM-Llama3-V 2.5 exhibits more trustworthy behaviors, achieving lower hallucination rates than GPT-4V-1106 on Object HalBench.
\item  {\textbf{Multilingual Support.}} Inspired by the findings from VisCPM~\citep{hu2023viscpm}, the integration of multilingual LLM significantly alleviates the heavy reliance on multimodal training data in low-resource languages. Based on the foundation, a high-quality multilingual multimodal instruction tuning helps MiniCPM-Llama3-V 2.5 generalize its multimodal capabilities to more than 30 languages.
\item  {\textbf{Efficient End-side Deployment.}} We systematically integrate a suite of end-side optimization techniques, encompassing quantization, memory optimization, compilation optimization and NPU acceleration, enabling efficient deployment on end-side devices.
\end{itemize}

We hope MiniCPM-V series can serve as an example for unveiling the potential of end-side MLLMs, and help draw more attention to promote the research in this direction.
Following Moore's Law for MLLM, we believe there will be increasingly powerful end-side MLLMs with reduced sizes, bringing efficient, safe, and trustworthy AI services on devices soon.

The contribution of this work is summarized as follows: (1) We introduce and open-source MiniCPM-V, a series of efficient end-side MLLMs achieving a good balance between performance and efficiency. (2) We investigate key techniques driving MLLMs towards the performance-efficiency balance at scale, unveiling the potential of these techniques. (3) We summarize the trend of MLLM development in its Moore's Law, and empirically instantiate the trend with representative examples of MiniCPM-V.

\begin{figure*}[!h]
\centering
\includegraphics[width=0.95\linewidth]{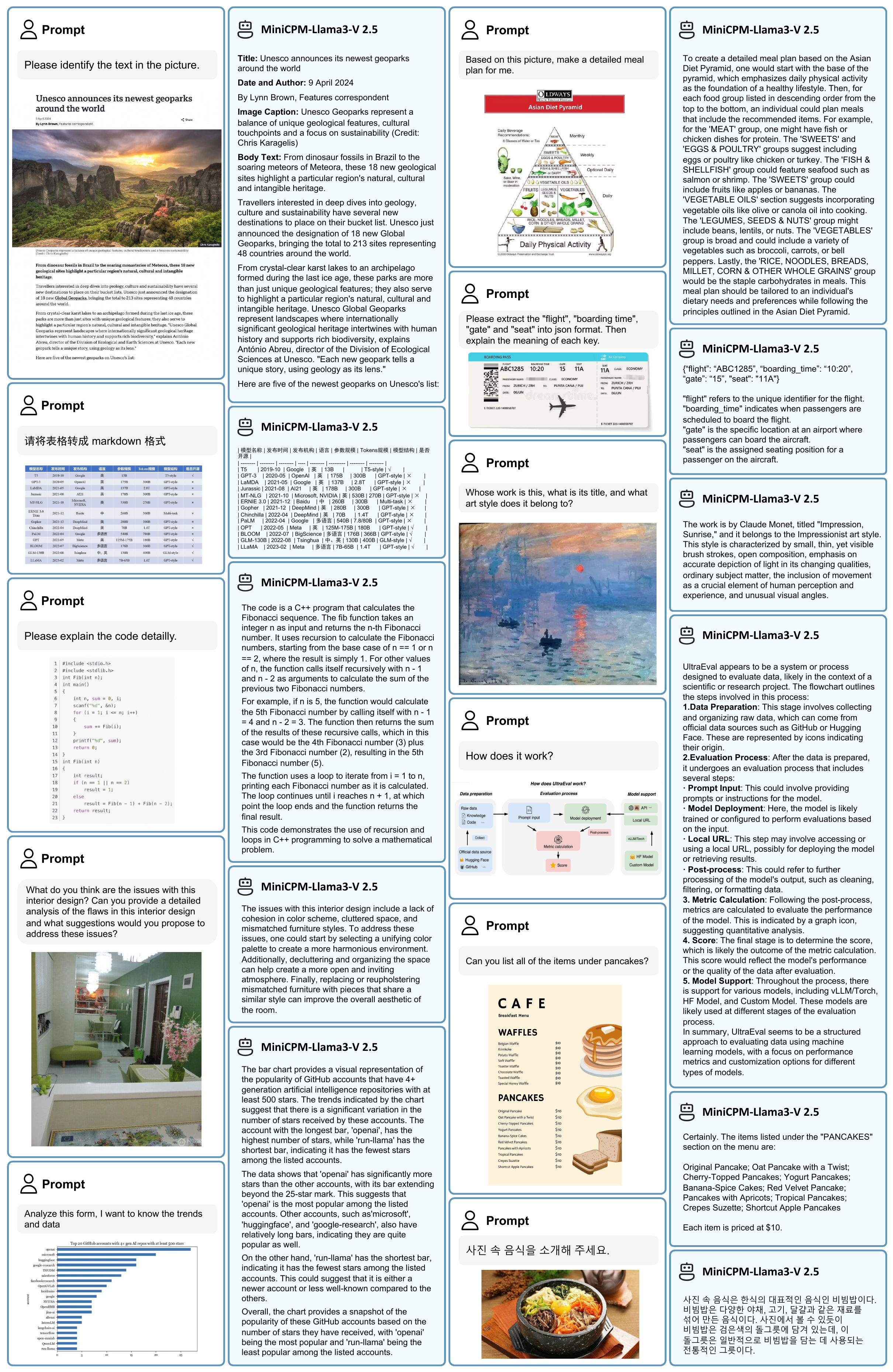}
\caption{Qualitative results of MiniCPM-Llama3-V 2.5 in reading text in images, converting tables to markdown, performing complex reasoning, and multilingual interaction, etc. See \href{https://github.com/OpenBMB/MiniCPM-V?tab=readme-ov-file\#examples-}{here} for the screen recordings of MiniCPM-Llama3-V 2.5 running on mobile phones.}
\vspace{-1mm}
\label{fig:case-1page}
\end{figure*}

\begin{figure*}[t]
\centering
\includegraphics[width=1.0\linewidth]{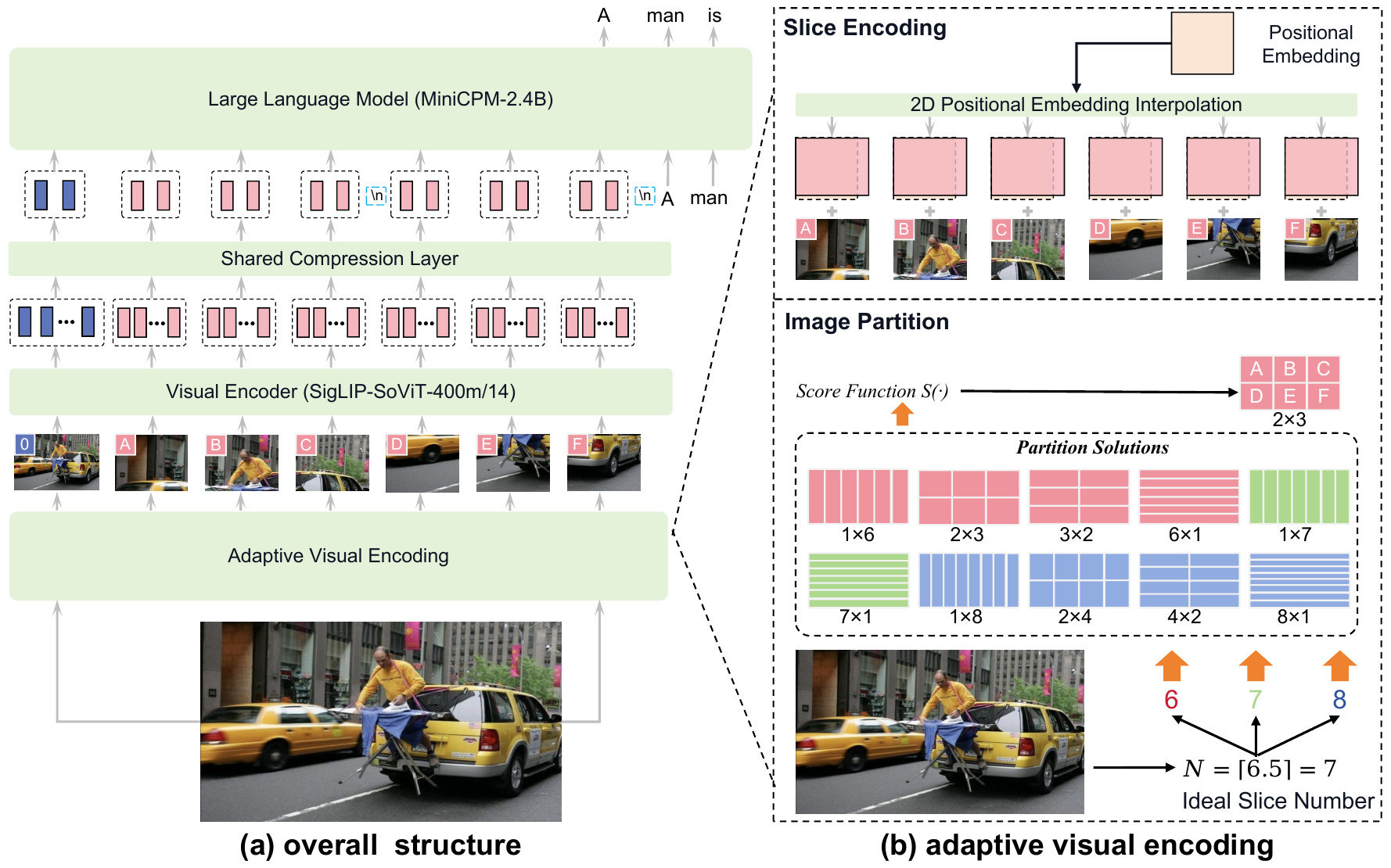}
\caption{Model architecture. (a) \textbf{Overall structure} presents the architecture of the model including the visual encoder, shared compression layer, and LLM. (b) \textbf{Adaptive visual encoding} deals with high-resolution inputs in various aspect ratios.}
\label{fig:framework}
\end{figure*}

\section{Related Works}
\paragraph{Multimodal Large Language Models.} The development of LLMs has significantly advanced the progress in MLLMs.
Flamingo~\citep{alayrac2022flamingo} first proposes to connect a pre-trained visual encoder with the Chinchilla 70B~\citep{hoffmann2022chinchilla} LLM and demonstrate the MLLM's zero-shot and few-shot ability across a series visual language tasks.
After the appearance of ChatGPT, many open-source models including BLIP-2~\citep{li2023blip2}, Kosmos-1~\citep{huang2023kosmos1}, MiniGPT-4~\citep{zhu2023minigpt}, LLaVA~\citep{liu2023llava}, and VPGTrans~\citep{zhang2024vpgtrans} are proposed.
Among them, most are built upon existing pre-trained LLMs like Llama~\citep{touvron2023Llama} and Vicuna~\citep{zheng2023vicuna}, while Kosmos-1 tries to train the LLM from scratch.
Later, researchers continue to extend the function scope of MLLMs and improve the visual perception capabilities.
Kosmos-2~\citep{peng2023kosmos2}, CogVLM~\citep{wang2023cogvlm}, Shikra~\citep{chen2023shikra}, and NExT-Chat~\citep{zhang2023nextchat} further incorporate the localization capabilities to the MLLMs with either pix2seq paradigm or connecting with detection/segmentation models.
Qwen-VL-Chat~\cite{qwenvl}, Yi-VL~\cite{ai2024yi}, DeepSeek-VL~\cite{lu2024deepseekvl}, InternVL~\cite{chen2024far} and Intern-XComposer~\cite{dong2024internlm} pay more attention to improving the models' capability with different techniques like high-resolution input, more training data, and better data ratio.

\paragraph{End-side Multimodal Large Language Models.} The huge number of parameters of MLLMs incurs prohibitively high computation costs in both training and deployment, greatly limiting the widespread applications. Recently, there has been a trend of building smaller LLMs with fewer parameters.
The representative models are Phi~\citep{Javaheripi2023Phi2}, Gemma~\cite{Banks2024Gemma}, MobileLLM~\citep{liu2024mobilellm}, MiniCPM~\citep{hu2024minicpm}, etc. The moderate size of these models makes them applicable on end-side devices such as personal computers and even mobile phones.
With optimized training strategies, end-side LLMs like MiniCPM 2B can achieve comparable performance with strong 7B models like Llama2-7B~\citep{touvron2023Llama2}. Similar trends have also been witnessed in MLLMs. For example, Mini-Gemini~\cite{li2024minigemini} and PaliGemma~\cite{beyer2024paligemma} are built based on Gemma 2B~\citep{Banks2024Gemma} and MobileVLM V2 is built based on MobileLlama~\citep{chu2023mobilevlm}.
However, the fewer-parameter nature of end-side MLLMs presents significant challenges to building a capable model. MiniCPM-V series aims to push forward the potential of end-side MLLMs by addressing the key bottleneck problems through careful designs in architecture, training, inference and deployment.

\section{Model Architecture}
In this section, we present the model architecture of MiniCPM-V, outlining the overall structure and the adaptive high-resolution visual encoding approach. The design philosophy of MiniCPM-V series is to achieve a good balance between performance and efficiency, a more practical objective for a broader scope of real-world applications, which is implemented in architecture design, training, inference, and deployment.

\subsection{Overall Structure}
The model comprises three key modules: the visual encoder, compression layer, and LLM. The input image is first encoded by a visual encoder, utilizing the adaptive visual encoding approach. Specifically, we employ SigLIP SoViT-400m/14~\cite{zhai2023sigmoid} as the visual encoder. The visual tokens are then compressed by the compression layer, which adopts a perceiver resampler structure with one layer cross-attention. Finally, the compressed visual tokens, along with the text input, are fed into the LLM for conditional text generation.

\subsection{Adaptive Visual Encoding}
\label{method:adp_venc}
Recently, there has been growing consensus on the fundamental role of visual encoding in MLLM performance~\citep{mckinzie2024mm1,lu2024deepseekvl}, especially for fine-grained capabilities such as OCR. For effectiveness, a good visual encoding strategy should both respect the raw aspect ratio of the input and preserve sufficient visual details (high resolution). For efficiency, the number of visual tokens from image encoding should be moderate to be affordable on end-side devices.
To this end, we take advantage of the adaptive visual encoding method proposed by LLaVA-UHD~\citep{xu2024llavauhd}.

\paragraph{Image Partition.}
To handle the high-resolution images with different aspect ratios, we divide images into slices, where each slice better matches ViT's pre-training setting in terms of resolution and aspect ratio.
Specifically, we first calculate the ideal number of slices based on the input image size. 
Given an image with resolution $(W_I, H_I)$ and a ViT pre-trained on images with resolution $(W_v, H_v)$, we calculate the ideal slice number $N=\lceil \frac{W_I\times H_I}{W_v\times H_v} \rceil$. 
Then, we choose the combination of rows $n$ and columns $m$ from the set $\mathbb{C}_N= \{(m, n)| m\times n = N, m\in \mathbb{N}, n\in \mathbb{N} \}$.
A good partition $(m, n)$ should result in slices that match well with ViT's pre-training setting.
To achieve this, we use a score function to evaluate each potential partition:
\begin{equation}
\small
    S(m, n)= -\left| \log \frac{W_I / m}{H_I / n} - \log \frac{W_v}{H_v}\right|.
\end{equation}

We select the partition with the highest score from all possible candidates:
\begin{equation}
\small
    m^*, n^* = \argmax_{(m,n)\in \bar{\mathbb{C}}} S(m, n),
\label{equ:partition}
\end{equation}
where $\bar{\mathbb{C}}$ is the possible $(m, n)$ combinations with the product $N$.
However, when $N$ is a prime number, the feasible solutions can be limited to $(N, 1)$ and $(1, N)$.
Therefore, we additionally introduce $\mathbb{C}_{N-1}$ and $\mathbb{C}_{N+1}$, and set $\bar{\mathbb{C}} = \mathbb{C}_{N-1} \cup \mathbb{C}_{N} \cup \mathbb{C}_{N+1}$. In practice, we set $N \textless 10$, supporting 1.8 million pixels (e.g., 1344 $\times$ 1344 resolution) at most during encoding. Although we can encompass more image slices for higher resolutions, we purposely impose this resolution upper-bound, since it already well covers most real-world application scenarios, and the benefit of further increasing encoding resolution is marginal considering the performance and overhead.

\paragraph{Slice Encoding.}
Although image partitioning can ensure a good match between the slices and the ViT pre-training setting, each slice's size is not precisely equal to $(W_v, H_v)$. To feed the slices into ViT, we first adjust each slice by resizing it proportionally so that the resultant area size matches ViT pre-training area size $W_v \times H_v$. This adjustment helps prevent a significant gap between the number of encoded patches and the ViT's pre-training setting.
Subsequently, we interpolate the ViT's position embeddings to adapt to the slice's ratio. This involves reshaping the ViT's 1D embedding $P_{1} \in \mathbb{R}^{Q \times l}$ back to its 2D format $P_{2} \in \mathbb{R}^{q \times q \times l}$, where the number of position embeddings $Q = q \times q$. Then, we interpolate $P_{2}$ to fit the size of each slice via 2D interpolation. We also include the original image as an additional slice to provide holistic information about the entire image.

\paragraph{Token Compression.}
After visual encoding, each slice is encoded into 1,024 tokens, where 10 slices can yield over 10k tokens collectively. To manage this high token count, we employ a compression module comprising of one-layer cross-attention and a moderate number of queries, with 2D positions informed~\cite{qwenvl}. In practice, the visual tokens of each slice are compressed into 64 queries for MiniCPM V1\&2 and 96 tokens for MiniCPM-Llama3-V 2.5 through this layer. Compared with other MLLMs with competitive performance, the significantly smaller number of visual tokens in MiniCPM-V series enables superior efficiency in terms of GPU memory consumption, inference speed, first-token latency and power consumption, making it more friendly to wider application scopes and communities.

\paragraph{Spatial Schema.}
To indicate each slice's position relative to the whole image, inspired by \cite{fuyu2023}, we additionally introduce a spatial schema.
We first wrap tokens of each slice by two special tokens \texttt{<slice>} and \texttt{<\textbackslash slice>}, and then employ a special token ``\textbackslash n'' to separate slices from different rows.

\section{Training}
The model training consists of 3 phases: the pre-training phase, the supervised fine-tuning phase, and the RLAIF-V phase. We will introduce the training recipe in the following sections.

\subsection{Pre-training}
In this phase, we utilize large-scale image-text pairs for MLLM pre-training.
The primary goal of this phase is to align the visual modules (i.e., visual encoder and compression layer) with the input space of the LLM and learn foundational multimodal knowledge. The pre-training phase is further divided into 3 stages.

\paragraph{Stage-1.} The role of stage-1 is to warm up the compression layer, primarily connecting the visual encoder and LLMs. (1) Trainable Modules. We randomly initialize the compression layer and train this module in stage-1, keeping other parameters frozen. The visual encoder's resolution is set to 224$\times$224, which is the same as the visual encoder's pre-training setting. (2) Data. To warm up the compression layer, we randomly select 200M data from the Image Captioning data in Table~\ref{tab:pretrain-data}. Data cleaning is performed to remove image-text pairs with poor correlation and ill-formatted text data, ensuring the data quality.

\begin{table}[h]
 \renewcommand\arraystretch{1.3}
    \centering
    \caption{Pre-training data. The pre-training data consists of image captioning and OCR data in  English and Chinese. LAION-2B-OCR is generated by applying OCR tools to LAION-2B images.}
    \label{tab:pretrain-data}
    \resizebox{0.8\textwidth}{!}{
    \begin{tabular}{c | l | c | c}
    \toprule
    \multicolumn{2}{c|}{Category} & Sources &  Size \\
    \midrule
    \multirow{4}{*}{Image Captioning} & \multirow{2}{*}{English} & COCO~\citep{lin2014mscoco}, VG~\citep{krishna2017vg}, CC3M~\citep{sharma2018cc3m}, CC12M~\citep{changpinyo2021cc12m} & \multirow{2}{*}{410M} \\
     & & LAION-COCO~\citep{schuhmann2022laion}, COYO~\citep{kakaobrain2022coyo-700m}, LAION-2B~\citep{schuhmann2022laion} & \\
     \cmidrule(lr){2-4} & \multirow{2}{*}{Chinese} & AIC~\citep{wu2017aic}, LAION-2B-Chinese~\citep{schuhmann2022laion}, WuKong~\citep{gu2022wukong} & \multirow{2}{*}{110M} \\
     & & Zero-Chinese~\citep{xie2022zero}, etc. &  \\
    \midrule
    \multirow{3}{*}{OCR+Knowledge} & \multirow{2}{*}{English} & WIT~\citep{srinivasan2021wit}, IDL~\citep{biten2022ocridl}, SynthText~\citep{gupta2016synthtext}, SynthDoG-en~\citep{kim2022synthdog} & \multirow{2}{*}{39M} \\
    & & SynthDoG-zh~\citep{kim2022synthdog}, ArxivCap~\citep{li2024arxivcap}, etc. & \\
    \cmidrule(lr){2-4} & Chinese &  WIT~\citep{srinivasan2021wit}, LAION-2B-OCR & 11M \\
    \bottomrule
    \end{tabular}}
\end{table}

\paragraph{Stage-2.}
After the warm-up training of the compression layer, the role of stage-2 is to extend the input resolution of the pre-trained visual encoder. (1) Trainable Modules. In stage-2, we extend the image resolution from 224$\times$224 to 448$\times$448.
The whole visual encoder is trained, leaving other parameters frozen. (2) Data. To extend the pre-trained resolution, we additionally select 200M data from the Image Captioning data in Table~\ref{tab:pretrain-data}.

\paragraph{Stage-3.}
After extending the primary input resolution of the visual encoder, we finally train the visual modules using the adaptive visual encoding strategy, which can further accommodate high-resolution inputs with any aspect ratio. (1) Trainable Modules. During the stage-3 training, both the compression layer and the visual encoder are trained to adapt to the language model embedding space. The LLM is kept frozen to avoid disruption from the relatively low-quality pre-training data. (2) Data. Different from the previous stages with only image captioning data, during the high-resolution pre-training stage, we additionally introduce OCR data to enhance the visual encoders' OCR capability.

\paragraph{Caption Rewriting.}
Image-text pairs sourced from the Web~\citep{schuhmann2022laion, kakaobrain2022coyo-700m} can suffer from quality issues in the caption data, including non-fluent content, grammatical errors, and duplicated words. Such low-quality data can lead to unstable training dynamics.
To address the issue, we introduce an auxiliary model for low-quality caption rewriting. The rewriting model takes the raw caption as input and is asked to convert it into a question-answer pair. The answer from this process is adopted as the updated caption. In practice, we leverage GPT-4~\citep{bubeck2023gpt4} to annotate a small number of seed samples, which are then used to fine-tune an LLM for the rewriting task.

\paragraph{Data Packing.}
Samples from different data sources usually have different lengths.
The high variance of sample lengths across batches will lead to inefficiency in memory usage and the risk of out-of-memory (OOM) errors.
To address the issue, we pack multiple samples into a single sequence with a fixed length. 
By truncating the last sample in the sequence, we ensure uniformity in sequence lengths, facilitating more consistent memory consumption and computational efficiency.
Meanwhile, we modify the position ids and attention masks to avoid interference between different samples. In our experiments, the data packing strategy can bring 2\textasciitilde3 times acceleration in the pre-training phase.

\paragraph{Multilingual Generalization.}  Multimodal capability across multiple languages is essential for serving users from broader communities. Traditional solutions involve extensive multimodal data collection and cleaning, and training for the target languages. Fortunately, recent findings from VisCPM~\citep{hu2023viscpm} have shown that the multimodal capabilities can be efficiently generalized across languages via a strong multilingual LLM pivot. This solution largely alleviates the heavy reliance on multimodal data in low-resource languages. In practice, we only pre-train our model on English and Chinese multimodal data, and then perform a lightweight but high-quality multilingual supervised fine-tuning to align to the target languages. Despite its simplicity, we find the resultant MiniCPM-Llama3-V 2.5 can achieve good performance in over 30 languages as compared with significantly larger MLLMs.

\begin{table}[t]
 \renewcommand\arraystretch{1.3}
    \centering
    \caption{SFT data for MiniCPM-V series. Part-1\&2 data are concatenated sequentially in the SFT phase. Part-1 focuses on bolstering basic recognition capabilities, while part-2 aims to enhance advanced capabilities in generating detailed responses and following human instructions.}
    \label{tab:sft-data}
    \resizebox{1\textwidth}{!}{
    \begin{tabular}{c | l | c | c}
    \toprule
    \multicolumn{2}{c|}{Category} & Sources &  Size \\
    \midrule
    \multirow{11}{*}{Part-1}& Short Caption & Flickr-30K~\citep{plummer2015flickr30k}, COCO~\citep{lin2014mscoco}  & 560K \\
     \cmidrule(lr){2-4} & \multirow{2}{*}{VQA} & FM-IQA~\citep{gao2015fmiqa}, VGQA~\citep{krishna2017vg}, IconQA~\citep{lu2021iconqa}, GQA~\citep{hudson2019gqa}, VQAv2~\citep{antol2015vqa} &  \multirow{2}{*}{1.4M} \\
     &  & CLEVR~\citep{johnson2017clevr}, VizWiz~\citep{gurari2018vizwiz}, Visual7W~\citep{zhu2016visual7w}, COCO-QA~\citep{ren2015cocoqa} & \\
     \cmidrule(lr){2-4} & Knowledge & OKVQA~\citep{marino2019okvqa}, A-OKVQA~\citep{schwenk2022aokvqa}, KVQA~\citep{shah2019kvqa}, ScienceQA~\citep{lu2022scienceqa}  & 60K \\
     \cmidrule(lr){2-4} & Grounding & RefCOCO~\citep{yu2016refcoco} & 570K \\
     \cmidrule(lr){2-4} & Reasoning & COMVINT~\citep{du2023comvint}, VCR~\citep{zellers2019vcr}, NLVR~\citep{suhr2017nlvr}, LRV~\citep{liu2023lrv} & 135K \\
     \cmidrule(lr){2-4} & Math & GeoQA~\citep{chen2021geoqa}, SMART-101~\citep{cherian2023smart101} & 125K \\
     \cmidrule(lr){2-4} & \multirow{3}{*}{OCR} & DocVQA~\citep{mathew2021docvqa}, TextVQA~\citep{singh2019textvqa}, OCR-VQA~\citep{mishra2019ocrvqa}, ST-VQA~\citep{biten2019stvqa}, VisualMRC~\citep{tanaka2021visualmrc}, DVQA~\citep{kafle2018dvqa} & \multirow{3}{*}{1.7M} \\
     &  &  FigureQA~\citep{kahou2017figureqa}, ChartQA~\citep{masry2022chartqa}, DeepForm~\citep{deepform}, TabFact~\citep{chen2019tabfact}, InfographicsVQA~\citep{mathew2022infographicvqa} & \\
     &  & Kleister Charity~\citep{stanislawek2021kleister}, WikiTableQuestions~\citep{pasupat2015wikitablequestions}, Real-CQA~\citep{ahmed2023realcqa}, AI2D~\citep{kembhavi2016ai2d}, etc. &  \\
     \cmidrule(lr){2-4} & Chat & FSVQA~\citep{shin2016fsvqa}, Visual-Dialog~\citep{das2017visualdialog} & 780K \\
     \midrule
      \multirow{7}{*}{Part-2} & Part-1 & sample from Part-1 data  & 400K \\
    \cmidrule(lr){2-4} & \multirow{2}{*}{OCR} & DocVQA, TextVQA, OCR-VQA, VisualMRC, ChartQA, AI2D  & \multirow{2}{*}{690K} \\
      &  & ArxivQA~\citep{li2024arxivqa}, LLaVAR~\citep{zhang2023llavar}, TextOCR-GPT4V~\citep{textocr-gpt4v}, etc. & \\
     \cmidrule(lr){2-4} & \multirow{2}{*}{Instruct} & SVIT~\citep{zhao2023svit}, LLaVA-Instruct-150K~\citep{liu2023llava}, UniMM-Chat~\citep{yu2023unimm}, ShareGPT4V~\citep{chen2023sharegpt4v} & \multirow{2}{*}{1.9M} \\
     &  & LVIS~\citep{gupta2019lvis}, ALLaVA~\citep{chen2024allava} & \\
     \cmidrule(lr){2-4} & \multirow{2}{*}{Text-Only} & Ultra-Chat~\citep{ding2023ultrachat}, Alpaca~\citep{alpaca}, ShareGPT~\citep{zheng2023vicuna}, BELLE~\citep{belle} & \multirow{2}{*}{-} \\
      & & OpenOrca~\citep{OpenOrca}, OpenHermes~\citep{OpenHermes}, In-House-MiniCPM-SFT & \\
    \bottomrule
    \end{tabular}}
\end{table}

\subsection{Supervised Fine-tuning}
After learning foundational capabilities from pre-training, we perform supervised fine-tuning (SFT) on high-quality visual question answering datasets to further learn knowledge and interaction capability from human annotations.

\paragraph{Trainable Modules.} Compared with the pre-training phase which mainly uses crawled data from the Web, the SFT phase mainly utilizes high-quality datasets annotated by either human lablers or strong models such as GPT-4. Therefore, we unlock all model parameters to better exploit the data and learn rich knowledge during SFT phase.

\paragraph{Data.} 
Recent works~\citep{hu2024minicpm,reid2024gemini} show that data near the end of training plays a more important role in shaping the models' capabilities and response styles.
We categorize the SFT data into two parts. Part-1 focuses on bolstering the models' basic recognition capabilities, while part-2 is tailored to enhance their capabilities in generating detailed responses and following human instructions.
Specifically, part-1 data consists of the traditional QA/captioning datasets with relatively short response lengths, which helps enhance the model's basic recognition capabilities.
In comparison, part-2 encompasses datasets featuring long responses with complex interactions, either in text or multimodal context.
During SFT, these two parts of data are concatenated and sequentially fed into the model.
For MiniCPM-Llama3-V 2.5, we integrate 2M data from the recent Cauldron dataset~\citep{idefics2} for multimodal knowledge augmentation, and 90K multilingual data over 36 languages for boosting the multilingual conversation capability.

\subsection{RLAIF-V}

\begin{figure*}[t]
\centering
\includegraphics[width=1.0\linewidth]{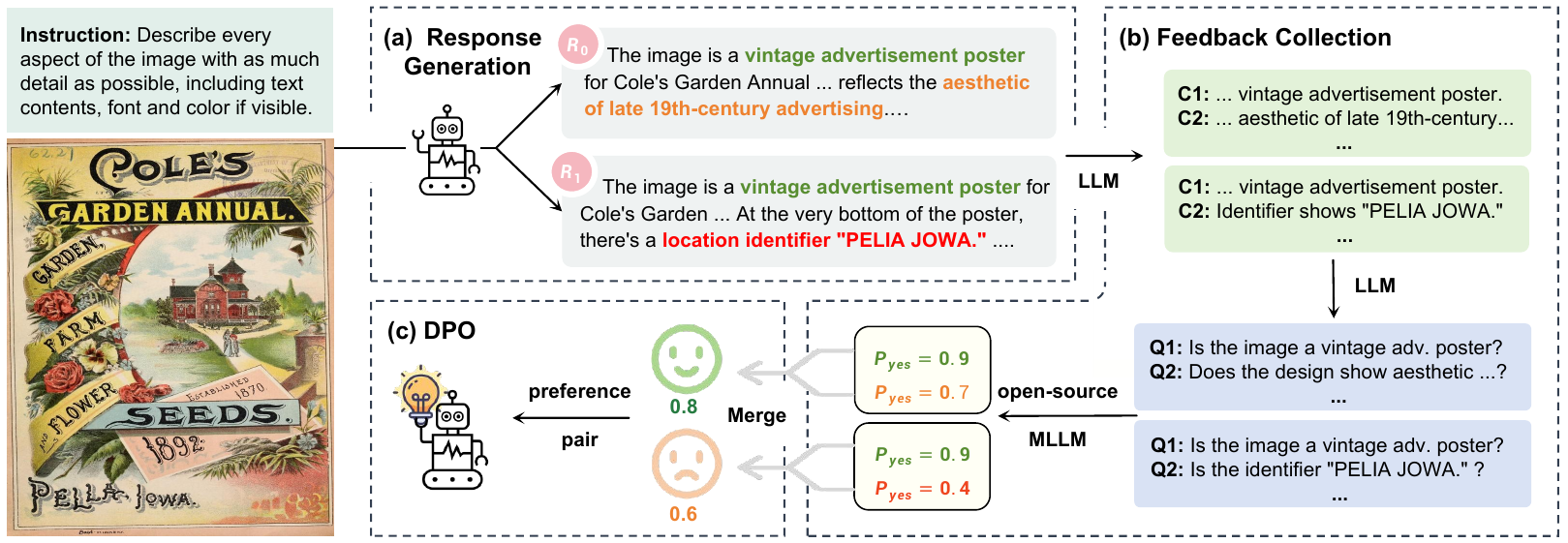}
\caption{RLAIF-V framework for hallucination reduction. (1) \textbf{Response generation} produces multiple responses for an instruction using the policy model. (2) \textbf{Feedback collection} evaluates the correctness of each response in a divide-and-conquer fashion. (3) \textbf{DPO} optimizes the model on the preference dataset.}
\label{fig:RLAIF-V}
\end{figure*}

MLLMs are typically prone to hallucination problems, generating responses that are not factually grounded in the input image~\cite{yu2024rlhf}. The issue greatly limits the wide application of MLLMs, especially in high-stakes scenarios, such as autonomous driving and assistance for visually impaired groups. To address the hallucination problem, we employ the recent RLAIF-V~\cite{yu2022rlaifv} approach (Fig.~\ref{fig:RLAIF-V}), where the key is to obtain scalable high-quality feedback from open-source models for direct preference optimization (DPO)~\cite{rafailov2023direct}.

\paragraph{Response Generation.} The first step of RLAIF-V is to generate multiple responses for a given instruction using the policy model.
Specifically, given a model $M$ waiting for alignment, we sample 10 responses $Y=\{y_1, y_2, \cdots, y_n\}$ from $M$ using sampling decoding with high temperatures.
There are several benefits of using the policy model $M$ for response generation: (1) Feedback collection and learning can better focus on trustworthiness, since different text styles from multiple MLLMs are avoided. (2) Feedback learning is more efficient since preference is directly collected on the distribution of the policy model.

\paragraph{Feedback Collection.} Collecting high-quality feedback from open-source MLLMs can be challenging due to their typically weaker capabilities compared with proprietary models. To address the issue, RLAIF-V uses a divide-and-conquer strategy for response scoring.
Specifically, each response $y_i$ is divided into atomic claims $C_i=\{c_1, c_2, \cdots, c_m\}$ using Llama-3 8B, where the correctness of atomic claims is much easier to evaluate.
Then, we verify the claims by converting each claim to a yes/no question and employing an open-source MLLM to score each claim. In practice, we adopt OmniLMM 12B for MiniCPM-V 2.0 scoring and LLaVA-NeXT-Yi 34B for MiniCPM-Llama3-V 2.5 scoring.
The final score $s_i$ of the response $y_i$ is given by  $-n_{rej}$, where $n_{rej}$ is the number of invalid atomic claims.

\paragraph{Direct Preference Optimization.} After collecting the high-quality AI feedback, we perform preference learning via DPO method.
The DPO algorithm requires training on preference pairs, where one sample $y_{w}$ is preferred to the other one $y_{l}$.
To compose the preference dataset, we randomly sample pairs from each response set $Y=\{y_1, y_2, \cdots, y_n\}$, and determine $(y_{w}, y_{l})$ based on their relative scores. Finally, we construct a preference dataset consisting of 6K preference pairs from 3K unique images for preference learning.

\section{End-side Deployment}

In this section, we investigate the deployment of MiniCPM-V on end-side devices. 
We first introduce the challenges, and then present the basic and advanced practices for end-side deployment.
Finally, we analyze and discuss the evaluation results across different devices.

\begin{figure*}[t]
\centering
\includegraphics[width=0.9\linewidth]{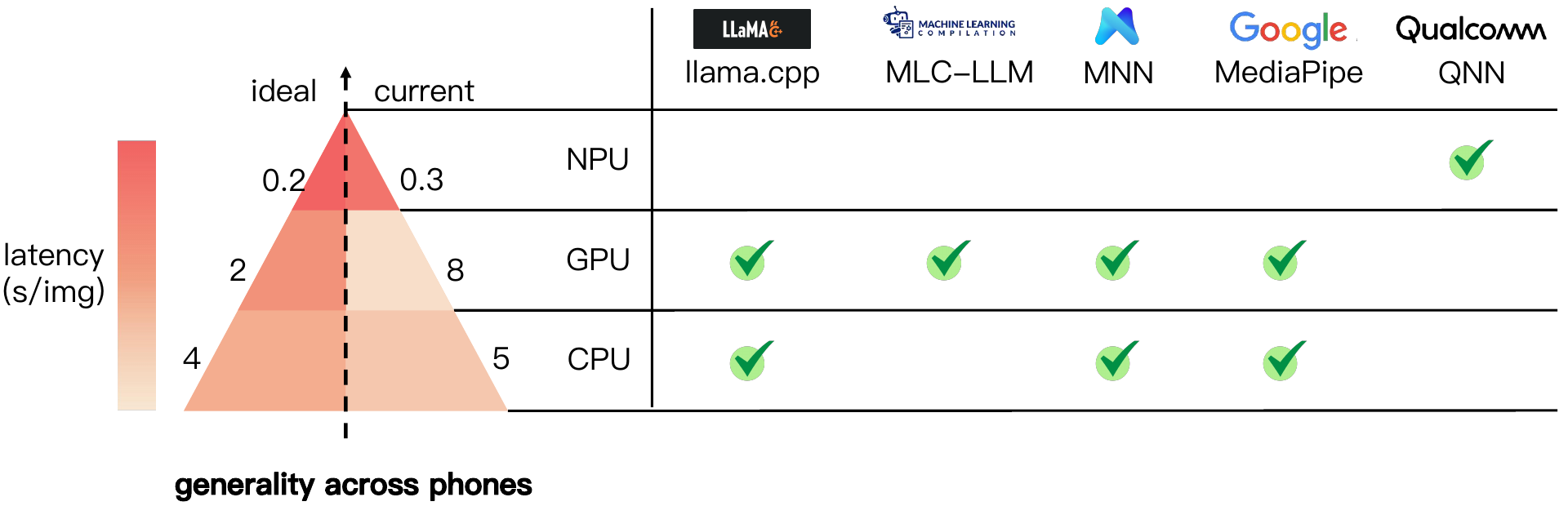}
\caption{An overview of end-side deployment. Current end-side chips for computation acceleration include CPU, GPU and NPU. We list their generality, (estimated) ideal and current performance, and related deployment frameworks.}
\label{fig:end_deploy_framework}
\end{figure*}

\subsection{Challenges}
End-side devices, such as smartphones and computers, often face resource limitations due to factors like heat dissipation, size constraints, and power consumption. We identify several key challenges of end-side deployment for MLLMs by comparing end-side devices with high-performance servers:

\paragraph{Memory Constraints.} High-performance servers typically boast extensive memory capacities, often exceeding 100GB or even 1TB. In contrast, the memory available on mobile phones typically ranges from 12GB to 16GB, which can be insufficient for MLLM deployment.

\paragraph{CPU/GPU Speed Restriction.} The overall processing speeds of CPUs in smartphones are notably slower. For instance, the Snapdragon 8 Gen3 features 8 CPU cores~\footnote{\url{https://docs.qualcomm.com/bundle/publicresource/87-71408-1_REV_E_Snapdragon_8_gen_3_Mobile_Platform_Product_Brief.pdf}}, whereas high-performance server like Intel Xeon Platinum 8580 has 60 CPU cores~\footnote{\url{https://www.intel.com/content/www/us/en/products/sku/237250/intel-xeon-platinum-8580-processor-300m-cache-2-00-ghz/specifications.html}}.
Similarly, mobile phone GPUs are not as powerful as server GPUs.
For example, Qualcomm Adreno 750 only has 6 TFLOPS, while NVIDIA 4090 can reach 83 TFLOPS. 

\subsection{Basic Practice}
To deploy the MLLM on end-side devices, we first employ quantization for reduced memory cost, and empirically investigate the deployment results on different frameworks.

\paragraph{Quantization.} Quantization is a widely used technique to reduce memory consumption.
The main idea of model quantization is to use a unified scaling factor to compress multiple weights into a narrower range, followed by discretization. This process is mathematically represented as: 
\begin{equation}
w'_i = \text{round}(\frac{w_i}{s}), \forall 1\le i\le n,
\end{equation}
where $w'$ denotes the quantized parameter and $s$ signifies the calculated scale factor. The $\text{round}$ function discretizes the quantized value.

For MiniCPM-Llama3-V 2.5, the fp16 version model typically demands 16\textasciitilde17G memory. We opt for the Q4\_K\_M mode 4-bit quantization strategy within GGML\footnote{\url{https://github.com/ggerganov/ggml}} framework. This reduces the memory requirement to around 5G, which is friendly to mobile phone usage.

\begin{figure*}[t]
\centering
\includegraphics[width=0.95\linewidth]{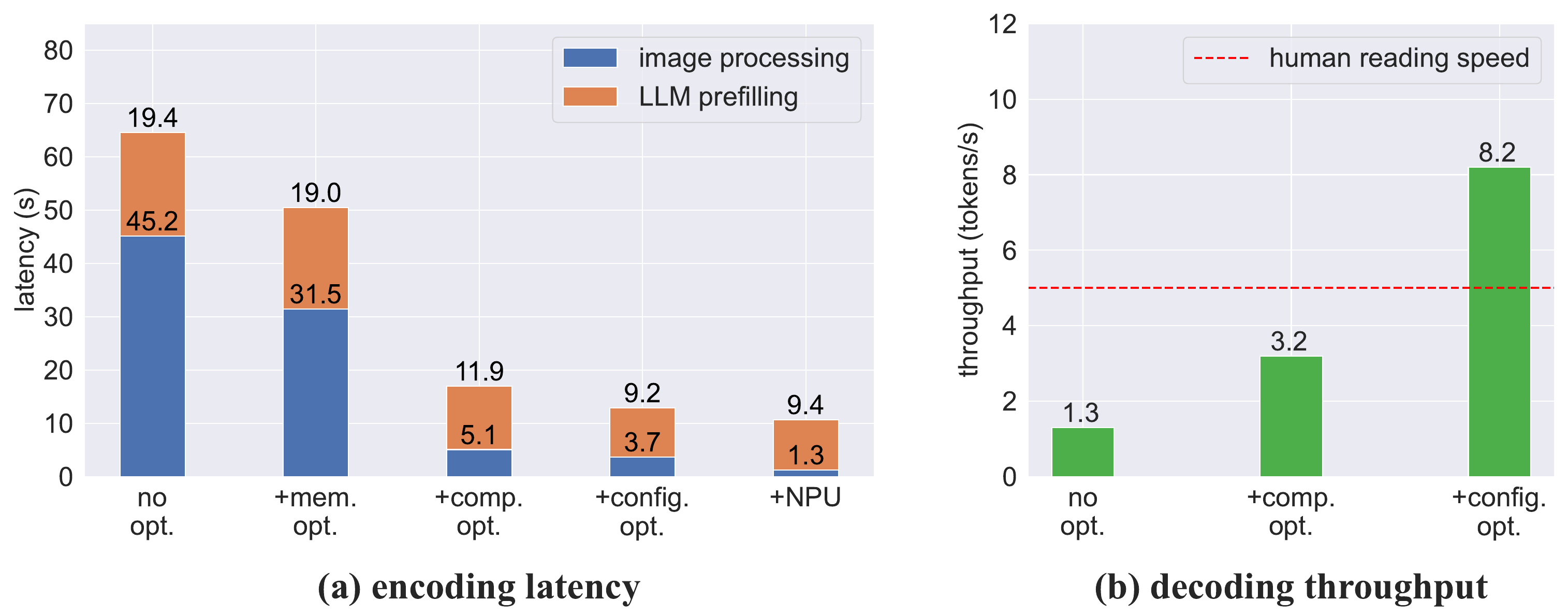}
\caption{Ablation study on the Xiaomi 14 Pro (Snapdragon 8 Gen 3). We show the influence of different techniques on the (a) encoding latency and (b) decoding throughput. \textbf{No opt.}: non-optimized, \textbf{mem. opt.}: memory usage optimization, \textbf{comp. opt.}: compilation optimization, \textbf{config. opt.}: configuration optimization, \textbf{NPU}: NPU acceleration. Note that the encoding latency includes both model loading time and encoding time, which differs from Fig.~\ref{fig:end_deploy_framework}'s encoding time only.} 
\label{fig:end_abl}
\end{figure*}

\paragraph{Deployment Framework.}
Several frameworks have been proposed for end-side deployment.
Illustrated in Fig.~\ref{fig:end_deploy_framework}, we make a thorough investigation of different frameworks for different chip types including CPU, GPU, and NPU.

Given the ubiquity of CPU usage across devices, we prioritize this chip type and opt for the llama.cpp~\citep{llamacpp} framework.
Combining quantization and llama.cpp on Xiaomi 14 Pro (Snapdragon 8 Gen 3), the model achieves a text encoding latency of 64.2s and a text decoding speed of 1.3 tokens/s (as depicted in Fig.~\ref{fig:end_abl}), which is still far from acceptable for users.

\subsection{Advanced Practice}
To enhance user experience, we investigate a series of advanced techniques including memory usage optimization, compilation optimization, configuration optimization, and NPU acceleration.

\paragraph{Memory Usage Optimization.}
Experimental results show that, without specific optimizations, image processing can be the bottleneck of the inference speed due to limited memory resources on mobile phones.
To address the issue, we explore memory usage optimization strategies.
Instead of loading both ViT and LLM simultaneously into memory, we adopt a sequential loading approach. Specifically, we first load ViT for visual encoding, followed by the LLM for visual and text token encoding. By releasing the large amount of memory occupied by LLM, we can prevent frequent paging (swapping in and out) during ViT encoding, thereby improving the program efficiency. This optimization technique, as illustrated in Fig.~\ref{fig:end_abl} (a), results in a notable reduction of image processing time from 45.2s to 31.5s. 

\paragraph{Compilation Optimization.}
We find that directly compiling the models on the target devices can significantly improve the encoding latency and the decoding throughput. This can be attributed to better consistency between the compilation and target device instruction set architecture.
As depicted in Fig.~\ref{fig:end_abl}, this optimization endeavor yields promising results. Encoding latency shows a notable reduction from 50.5s to 17.0s, while decoding throughput experiences a significant boost from 1.3 tokens/s to 3.2 tokens/s.

\paragraph{Configuration Optimization.} We observe that a single default configuration of the llama.cpp framework may not be optimal for diverse end-side devices.
To maximize the inference speed, we devise an automatic parameter search algorithm that dynamically determines the most suitable configurations (e.g., computation allocation on different CPU cores).
Through configuration optimization, we can achieve good improvements. Specifically, decoding throughput surged from 3.2 tokens/s to an impressive 8.2 tokens/s, surpassing the typical human reading speed. 

\paragraph{NPU Acceleration.}
The above techniques are mostly tailored for CPU deployment. Another promising avenue involves leveraging alternative chip types such as GPUs and NPUs.
Despite the potential of GPU, we find in our experiments that current frameworks for mobile phone GPU are not optimized or compatible enough to exceed the results on CPU.
As an alternative, we turn to NPUs (Neural Processing Units), which represent a novel class of specialized hardware introduced in recent years, specifically designed for accelerating AI applications. Some smartphones are already equipped with NPUs, which are recognized as better suited for addressing computation bottlenecks.

In practice, we primarily leverage NPUs to accelerate visual encoding.
Specifically, we replace the backend framework of ViT to QNN, while retaining the llama.cpp backend for the LLM component.
For mobile phones equipped with Qualcomm NPUs, this optimization results in a notable reduction in visual encoding time, decreasing from 3.7s to 1.3s, as illustrated in Fig.~\ref{fig:end_abl} (a).

\begin{figure*}[t]
\centering
\includegraphics[width=0.8\linewidth]{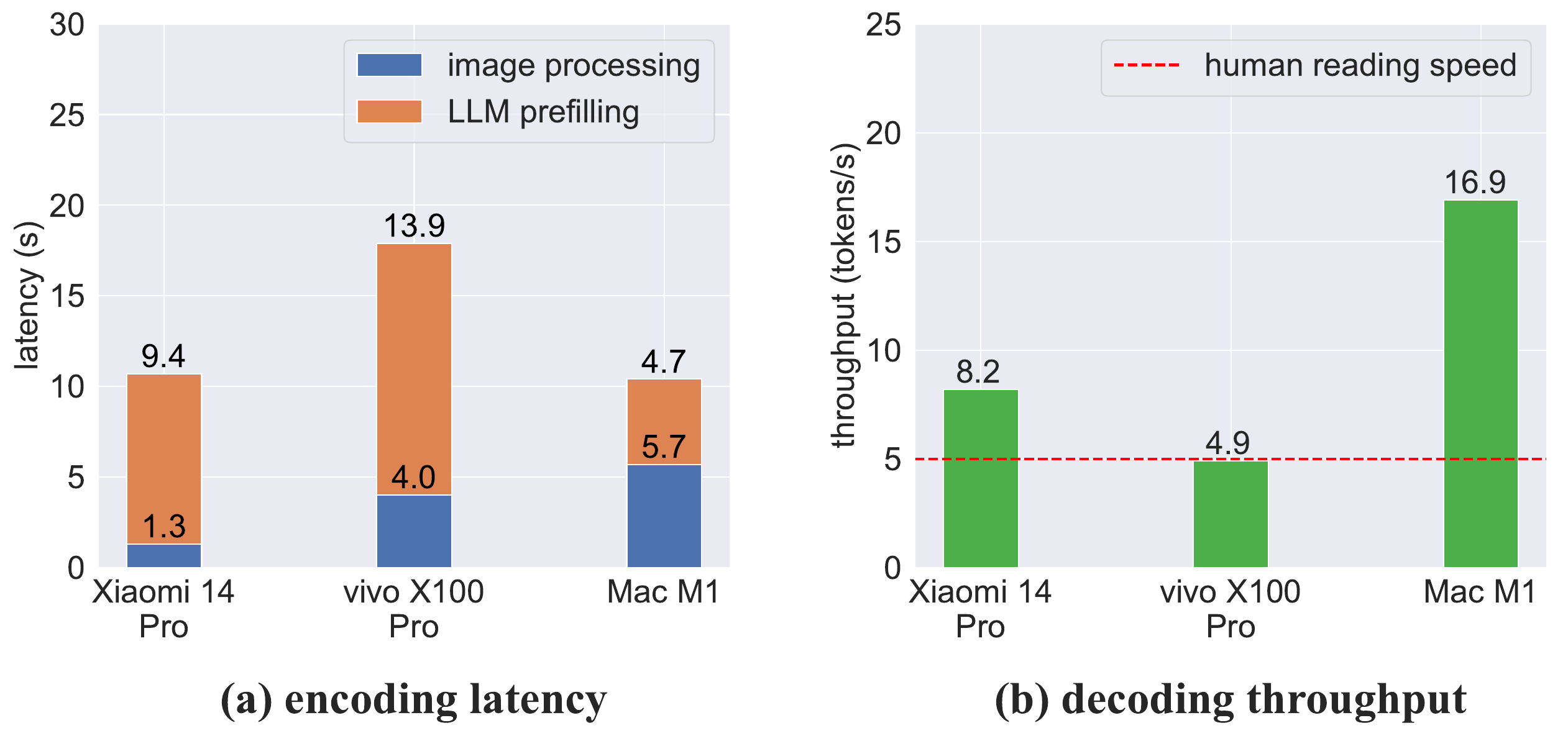}
\caption{Results on different end-side devices. We show the (a) encoding latency and (b) decoding throughput across different device types. Xiaomi 14 Pro is the only device with NPU.}
\label{fig:end_rst}
\end{figure*}
\subsection{Results.}
\paragraph{Analysis.} For a comprehensive assessment of MiniCPM-Llama3-V 2.5's performance across various end-side devices, we present test results on Xiaomi 14 Pro (Snapdragon 8 Gen 3), vivo X00 Pro (Mediatek Dimensity 9300), and Macbook Pro (M1) in Fig.~\ref{fig:end_rst}.
Thanks to the deployment optimization techniques, MiniCPM-Llama3-V 2.5 can operate efficiently on both mobile phones and personal computers, delivering acceptable latency and throughput.
For instance, leveraging NPU on Xiaomi 14 Pro enables it to achieve a similar encoding speed as the Mac M1. Furthermore, nearly all devices exhibit comparable or higher throughput compared with human reading speed.

\paragraph{Discussion.} Upon analyzing Fig.~\ref{fig:end_rst}, it becomes evident that the current computation bottleneck primarily stems from LLM prefilling, which mainly involves encoding image and text tokens for LLM inference. Promising research directions involve developing more efficient visual encoding methods with fewer visual tokens, and better leveraging GPU/NPU acceleration for LLM encoding.
With increasing attention to end-side MLLMs and the rapid advancement of GPU/NPU acceleration techniques, we believe that real-time interaction with end-side MLLMs can be reached soon.

\section{Experiments}
In this section, we perform a comprehensive evaluation of MiniCPM-V series. 

\subsection{MiniCPM-V Series}
We have released 3 models in the MiniCPM-V series, including MiniCPM-V 1.0, MiniCPM-V 2.0, and MiniCPM-Llama3-V 2.5.
As shown in Table~\ref{tab:minicpmv-series}, MiniCPM-V 1.0 is trained with the pre-training stage1\&2 and SFT without using the adaptive visual encoding and RLAIF-V.
For MiniCPM-V 2.0, we include all of the training stages and the adaptive visual encoding strategy to further improve performance.
In MiniCPM-Llama3-V 2.5, Llama3-Instruct 8B is adopted as the base LLM.

\begin{table}[h]
    \centering
    \caption{The MiniCPM-V series, with key components and configurations. AR.: aspect ratio.}
    \label{tab:minicpmv-series}
    \resizebox{\textwidth}{!}{
    \begin{tabular}{l l l l  l l l l}
    \toprule
    Model & Base LLM & Resolution & AR. & Pre-training & SFT & Alignment \\
    \midrule
    MiniCPM-V 1.0 & MiniCPM 2B & 0.2M pixel (i.e., 448 $\times$ 448) & Fixed  & stage-1\&2 & part1+2 & No \\
    MiniCPM-V 2.0 & MiniCPM 2B & 1.8M pixel (e.g., 1344 $\times$ 1344) & Any  & stage-1\&2\&3 & part1+2 & RLHF-V \\
    MiniCPM-Llama3-V 2.5 & Llama3-Instruct 8B & 1.8M pixel (e.g., 1344 $\times$ 1344) & Any & stage-1\&2\&3 & part1+2 & RLAIF-V \\    
    \bottomrule
    \end{tabular}}
\end{table}

\subsection{Experiment Settings}
\paragraph{Benchmarks.} We perform a comprehensive evaluation on popular benchmarks covering visual question answering, multimodal conversation, knowledge and reasoning, OCR, and hallucination. (1) \textbf{General benchmarks.} We adopt OpenCompass~\citep{2023opencompass} as the general evaluation indicator, which is a comprehensive collection over 11 popular multimodal benchmarks, including MME~\citep{fu2023mme}, MMBench~\citep{liu2023mmbench}, MMMU~\citep{yue2023mmmu}, MathVista~\citep{lu2023mathvista}, LLaVA Bench~\citep{liu2023llava}, etc. We also report the results on RealWorldQA for real-world spatial understanding capabilities. (2) \textbf{OCR benchmarks.} We adopt three widely used benchmarks for OCR capability evaluation, including including OCRBench~\citep{liu2023ocrbench}, TextVQA~\citep{singh2019textvqa} and DocVQA~\citep{mathew2021docvqa}. (3) \textbf{Hallucination benchmarks.} We also include Object HalBench~\citep{rohrbach2018objhalbench,yu2024rlhf} to evaluate the trustworthiness of the models.

\paragraph{Baselines.} We compare with strong baselines in different series: For open-source models, we compare with strong models including Yi-VL-6B/34B~\citep{ai2024yi}, Qwen-VL-Chat~\citep{qwenvl}, DeepSeek-VL-7B~\citep{lu2024deepseekvl}, TextMonkey~\citep{liu2024textmonkey}, CogVLM-Chat-17B~\citep{wang2023cogvlm}, CogVLM2-Llama3-19B~\citep{wang2023cogvlm},  Idefics2-8B~\citep{idefics2}, Bunny-Llama-3-8B~\citep{he2024bunnyllama3}, XTuner-Llama-3-8B-v1.1~\citep{2023xtunerllama3}, LLaVA-NeXT-Llama-3-8B~\citep{li2024llavanext-strong}, Cambrian-8B/34B~\cite{tong2024cambrian}, LLaVA-NeXT-Yi-34B~\cite{liu2024llavanext}, DeepSeek-VL-1.3B~\citep{lu2024deepseekvl}, MobileVLM V2~\citep{chu2023mobilevlm}, Mini-Gemini~\citep{li2024minigemini} and Phi-3-Vision-128k-instruct~\citep{abdin2024phi3}. For proprietary models, we compare with GPT-4V-1106~\citep{achiam2023gpt4}, Gemini-Pro~\citep{reid2024gemini} and Claude 3 Opus~\cite{claude2024}.

\begin{table}[t]
    \centering
    \caption{Experimental results on general multimodal benchmarks. RW QA: RealWorldQA, Obj HalBench (Res./Men.) : Object HalBench with response/mention-level hallucination rates, *: our tested results with official checkpoints. The best open-source results are highlighted in \textbf{bold}.}
    \label{tab:cmp-general}
    \resizebox{\textwidth}{!}{
    \begin{tabular}{l l l l l l l l l l l}
    \toprule
    \multirow{2}{*}{Model} & \multirow{2}{*}{Size} & Open- & \multirow{2}{*}{MME} & MMB & MMB & MMMU & Math- & LLaVA & RW & Obj HalBench \\
     &  & Compass &  & test (en) & test (cn) & val & Vista & Bench & QA & (Res./Men.) $\downarrow$\\
    \midrule
    \textbf{Proprietary} &\\
    GPT-4V (2023.11.06) & - & 63.5 & 1771.5 & 77.0 & 74.4 & 53.8 & 47.8 & 93.1 & 63.0 & 13.6 / 7.3*  \\
    Gemini Pro & - & 62.9 & 2148.9 & 73.6 & 74.3 & 48.9 & 45.8 & 79.9 & 60.4 & - \\
    Claude 3 Opus & - & 57.7 & 1586.8 & 63.3 & 59.2 & 54.9 & 45.8 & 73.9 & 48.4 & - \\
    \midrule
    \textbf{Open-source} &\\
    DeepSeek-VL-1.3B & 1.7B & 46.2 & 1531.6 & 66.4 & 62.9 & 33.8 & 29.4 & 51.1 & 49.7 & 16.7 / 9.6* \\
    Mini-Gemini & 2.2B & - & 1653.0 & - & - & 31.7 & - & - & - & - \\
    Yi-VL-6B & 6.7B & 48.9 & 1915.1 & 68.4 & 66.6 & 40.3 & 28.8 & 51.9 & 53.5 & 19.4 / 11.7* \\
    Qwen-VL-Chat & 9.6B & 51.6 & 1860.0 & 61.8 & 56.3 & 37.0 & 33.8 & 67.7 & 49.3 & 43.8 / 20.0* \\
    Yi-VL-34B & 34B & 52.2 & \textbf{2050.2} & 	72.4 &  70.7 & 45.1 & 30.7 & 62.3 & 54.8 & 20.7 / 14.0* \\
    Phi-3-vision-128k-instruct & 4.2B & - & - & - & - & 40.4 & 44.5 & 64.2* & 58.8* & - \\
    XTuner-Llama-3-8B-v1.1 & 8.4B & 53.3  & 1818.0 & 71.7 & 63.2 & 39.2 & 40.0 & 69.2 & - & - \\
    CogVLM-Chat & 17B & 54.2 & 1736.6 &  65.8& 55.9 & 37.3 & 34.7 & 73.9 & 60.3 & 26.4 / 12.6* \\
    Bunny-Llama-3-8B & 8.4B & 54.3 & 1920.3 & 77.0 & 73.9 & 41.3 & 31.5 & 61.2 & 58.8 & - \\
    DeepSeek-VL-7B & 7.3B & 54.6 & 1765.4 & 73.8 & 	71.4 & 38.3 & 36.8 & 77.8 & 54.2 & 11.4 / 6.5* \\
    LLaVA-NeXT-Llama3-8B & 8.4B & - & 1971.5 & - & - & 41.7 & -  & 80.1 & 60.0 & - \\
    Idefics2 & 8.0B & 57.2 & 1847.6 & 75.7 & 68.6 & 45.2 & 52.2 & 49.1 & 60.7 & - \\
    Cambrian-8B & 8.3B & 58.8 & 1802.9 & 74.6 & 67.9 & 41.8 & 47.0  & 71.0 & 60.0 & - \\
    CogVLM2-19B-Chat & 19B & 62.3 & 1869.5 & 73.9 & 69.8 & 42.6 & 38.6  & 83.0 & 62.9 & - \\
    LLaVA-NeXT-Yi-34B & 34B & 62.7 & 2006.5 & \textbf{81.1} & 79.0 & 48.8 & 40.4  & 81.8 & 66.0 & - \\
    Cambrian-34B & 34B & 64.9 & 2049.9 & 80.4 & \textbf{79.2} & \textbf{50.4} & 50.3 & 82.0 & \textbf{67.1} & - \\
    \midrule
    \rowcolor[RGB]{230, 242, 255} MiniCPM-V 1.0 & 2.8B & 47.5 & 1650.2 & 64.1 & 62.6 & 38.3 & 28.9 & 51.3 & 51.2 & 21.6 / 11.5 \\
    \rowcolor[RGB]{230, 242, 255} MiniCPM-V 2.0 & 2.8B & 54.5 & 1808.6 & 69.1 & 66.5 & 38.2 & 38.7 & 69.2 & 55.8 & 14.5 / 7.8 \\
    \rowcolor[RGB]{230, 242, 255} MiniCPM-Llama3-V 2.5 & 8.5B & \textbf{65.1} & 2024.6 & 77.2 & 74.2 & 45.8 & \textbf{54.3} & \textbf{86.7} & 63.5 & \textbf{10.3} / \textbf{5.0} \\
    \bottomrule
    \end{tabular}}
\end{table}

\begin{table}[t]
    \centering
    \caption{Results on OCR benchmarks. *: our tested results with official checkpoints. The best results are marked in \textbf{bold}.} 
    \label{tab:cmp-ocr}
    \resizebox{0.7\textwidth}{!}{
    \begin{tabular}{l l l l l}
    \toprule
    Model &Size & OCRBench & TextVQA val & DocVQA test  \\
    \midrule
    \textbf{Proprietary} &\\
    Gemini Pro & - & 680 & 74.6 & 88.1 \\
    GPT-4V (2023.11.06) & - & 645 & 78.0 & 88.4 \\
    \midrule
    \textbf{Open-source} &\\
    Yi-VL-6B & 6.7B & 290 & 45.5* & 17.1* \\
    Yi-VL-34B & 34B & 290 & 43.4* & 16.9* \\
    Mini-Gemini & 2.2B & - & 56.2 & 34.2* \\
    MobileVLM V2 & 3.1B & - & 57.5 & 19.4* \\
    DeepSeek-VL-1.3B & 1.7B & 413 & 58.4* & 37.9* \\
    Qwen-VL-Chat & 9.6B & 488 & 61.5 & 62.6 \\
    DeepSeek-VL-7B & 7.3B & 435 & 64.7* & 47.0* \\
    CogVLM-Chat & 17.4B & 590 & 70.4 & 33.3* \\
    TextMonkey & 9.7B & 558 & 64.3 & 66.7 \\
    Idefics2 & 8.0B & - & 73.0 & 74.0 \\
    Phi-3-vision-128k-instruct & 4.2B & 639* & 70.9 & - \\
    \midrule
    \rowcolor[RGB]{230, 242, 255} MiniCPM-V 1.0 & 2.8B & 366 & 60.6 & 38.2 \\
    \rowcolor[RGB]{230, 242, 255} MiniCPM-V 2.0 & 2.8B & 605 & 74.1 & 71.9 \\
    \rowcolor[RGB]{230, 242, 255} MiniCPM-Llama3-V 2.5 & 8.5B & \textbf{725} & \textbf{76.6} & \textbf{84.8} \\
    \bottomrule
    \end{tabular}}
\end{table}

\begin{figure*}[t]
\centering
\includegraphics[width=0.95\linewidth]{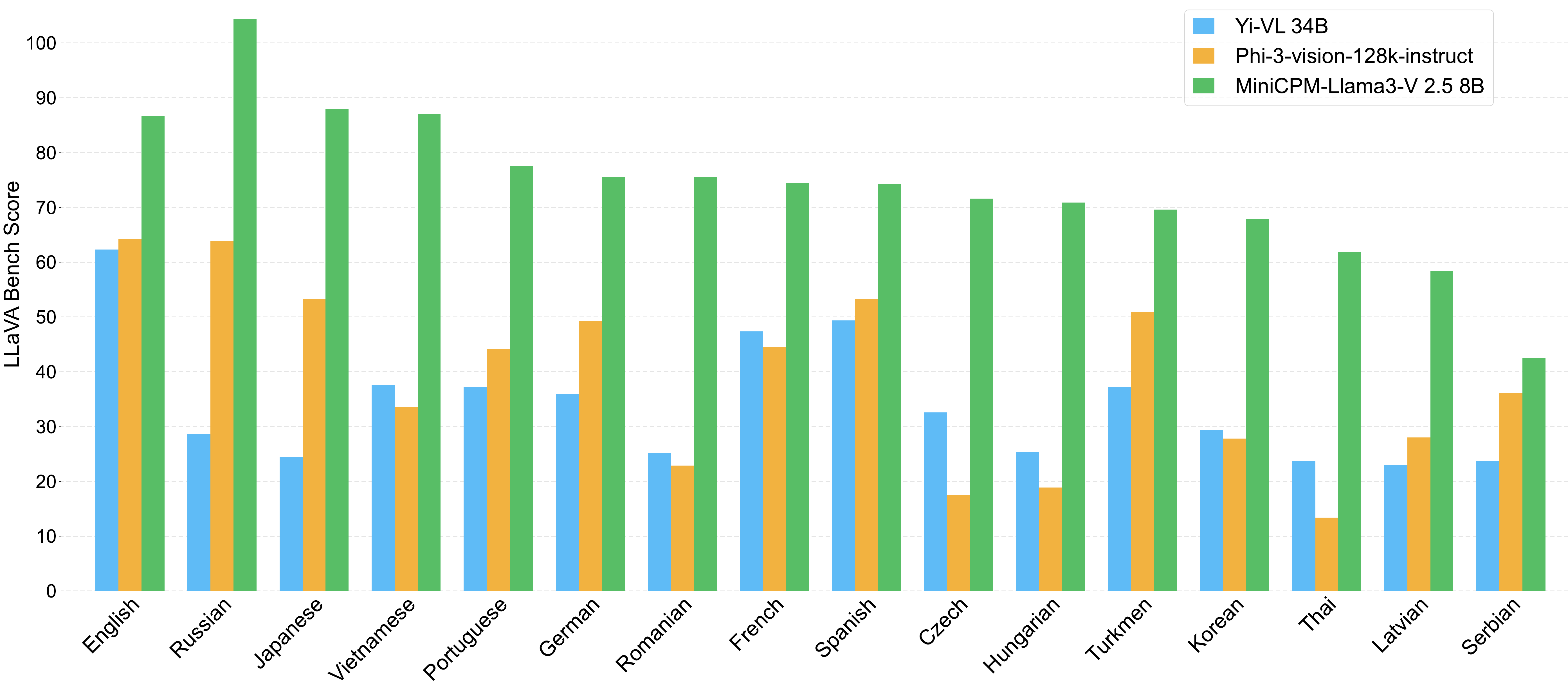}
\caption{Experimental results on multilingual multimodal interaction. We translate LLaVA Bench reference response into different languages, and adopt GPT-4-Turbo for evaluation. Here we randomly select 15 languages for visualization.}
\label{fig:multilingual}
\end{figure*}

\subsection{Experimental Results}
\paragraph{Main Results on General Multimodal Benchmarks.} From the experimental results in Table~\ref{tab:cmp-general}, we have the following observations: 
(1) MiniCPM-Llama3-V 2.5 outperforms strong open-source models by a notable margin. For instance, MiniCPM-Llama3-V 2.5 surpasses the recent strong Idefics2-8B by 7.9 points on the OpenCompass benchmark, with similar model sizes. It also achieves better results than significantly larger models such as Cambrian-34B, LLaVA-NeXT-Yi-34B, Yi-VL-34B and CogVLM2-Llama3-19B.
(2) Compared with powerful proprietary models, such as GPT-4V-1106 and Gemini Pro, MiniCPM-Llama3-V 2.5 achieves better performance on the OpenCompass benchmark with significantly fewer parameters.
In addition, MiniCPM-Llama3-V 2.5 also achieves lower hallucination rates than GPT-4V-1106 on Object HalBench, indicating its trustworthiness for real-world applications.
(3) The smaller MiniCPM-V 2.0 with 2B parameters achieves significantly better performance compared with other 2B\textasciitilde3B models, and is even comparable with Llama3-based 8B MLLMs such as Bunny-Llama-3-8B. In summary, the results show that MiniCPM-V series achieves a good balance between performance and efficiency, making it more friendly for broader communities and applications.

\paragraph{Results on OCR Benchmarks.}
MiniCPM-V models also show strong OCR capabilities, including scene-text, document and screenshot understanding.
As shown in Table~\ref{tab:cmp-ocr}, MiniCPM-Llama3-V 2.5 outperforms open-source MLLMs ranging 1.7B\textasciitilde34B on OCRBench, TextVQA, and DocVQA, and even performs comparably to proprietary models such as GPT-4V-1106 and Gemini Pro.

\paragraph{Multilingual Multimodal Capability.} Based on the multilingual multimodal generalization approach from VisCPM, MiniCPM-Llama3-V 2.5 extends its multimodal capability to over 30 languages. As shown in Fig.~\ref{fig:multilingual}, MiniCPM-Llama3-V 2.5 can outperform Yi-VL 34B and Phi-3-vision-128k-instruct on the multilingual LLaVA Bench. The promising multilingual multimodal capability makes MiniCPM-Llama3-V 2.5 useful in serving larger groups with various languages.

\paragraph{Comparison with Other Llama-3 based Models.} From experimental results in Table~\ref{tab:cmp-general}, we can observe that: (1) MiniCPM-Llama3-V 2.5 outperforms other Llama-3 based models by a large margin.
For example, compared with the strong LLaVA-NeXT-Llama-3-8B, MiniCPM-Llama3-V 2.5 consistently achieves better results on all benchmarks.
(2) Moreover, it is worth noting that MiniCPM-Llama3-V 2.5 requires significantly less inference computation. For example, the visual token number range of MiniCPM-Llama3-V 2.5 is (96, 960), which is lower than LLaVA-NeXT-Llama-3-8B's (1728, 2880). This can be important especially for real-world end-side applications in terms of inference speed, first-token latency, memory usage, and power consumption.

\subsection{Ablation Study}
We perform an ablation study on components of MiniCPM-Llama3-V 2.5, including RLAIF-V and multilingual training.

\begin{table}[t]
    \centering
    \caption{The influence of RLAIF-V. The MLLM for the ablation is MiniCPM-Llama3-V 2.5.}
    \label{tab:abl-RLAIF-V}
    \resizebox{0.9\textwidth}{!}{
    \begin{tabular}{l l l l l l l l l}
    \toprule
    \multirow{2}{*}{Method}  & Open- & \multirow{2}{*}{MME} & MMB & MMB & MMMU & Math- & LLaVA &  Object \\
     & Compass &  & dev(en) & dev(zh) & val & Vista & Bench &  HalBench \\
    \midrule
    w/o RLAIF-V &  64.5 & 2039.8 & 77.7 & 73.5 & 46.2 & 54.1 & 85.4 & 86.9 / 93.6\\
    w RLAIF-V &  65.1 & 2024.6 & 77.2 & 74.2 & 45.8 & 54.3 & 86.7  & 89.7 / 95.0 \\
    \bottomrule
    \end{tabular}}
\end{table}

\begin{table}[t]
    \centering
    \caption{The influence of multilingual generalization. We use 90k multilingual data ($< 0.5\%$ SFT data) for post SFT training, and show the performance changes below. ML: multilingual training.}
    \label{tab:abl-ml}
    \resizebox{\textwidth}{!}{
    \begin{tabular}{l l l l l l l l l l}
    \toprule
    Method  & French & German & Portuguese & Spanish & Czech & Hungarian & Japanese & Korean & Thai    \\
    \midrule
    w/o ML & 46.4 & 22.8 & 53.0 & 29.0 & 26.5 & 20.6 & 13.8 & 13.7 & 14.4  \\
    w ML & 72.7 & 76.5 & 83.8 & 73.9 & 71.6 & 70.9 & 88.0  & 67.9 & 61.9 \\
    \bottomrule
    \end{tabular}}
\end{table}

\paragraph{Influence of RLAIF-V.}
From the results in Table~\ref{tab:abl-RLAIF-V}, we can observe that RLAIF-V effectively reduces the hallucination rates of the base model on both response level and mention level. This makes the model more trustworthy in behaviors. Importantly, the hallucination reduction does not sacrifice the general capabilities. In contrast, RLAIF-V further improves the overall performance on OpenCompass by 0.6 points on an average of 11 benchmarks.

\paragraph{Multilingual Generalization.} We investigate the necessity and effectiveness of the multilingual generalization technique.
As shown in Table~\ref{fig:multilingual}, we can see over 25 point improvement in all languages when using less than 0.5\% multilingual SFT data. The results show that the multilingual generalization method can effectively improve multilingual capability with good data and computation efficiency. In addition, we notice that the performance improvement is uneven across languages.
We hypothesize that the improvement extent might be related to multiple factors like the base LLM's ability of the given language. We leave more systematical exploration for future works.

\subsection{Case Study}
We provide a more intuitive understanding of MiniCPM-Llama3-V 2.5 capabilities in the case study.

\begin{figure*}[!t]
\centering
\includegraphics[width=1.0\linewidth]{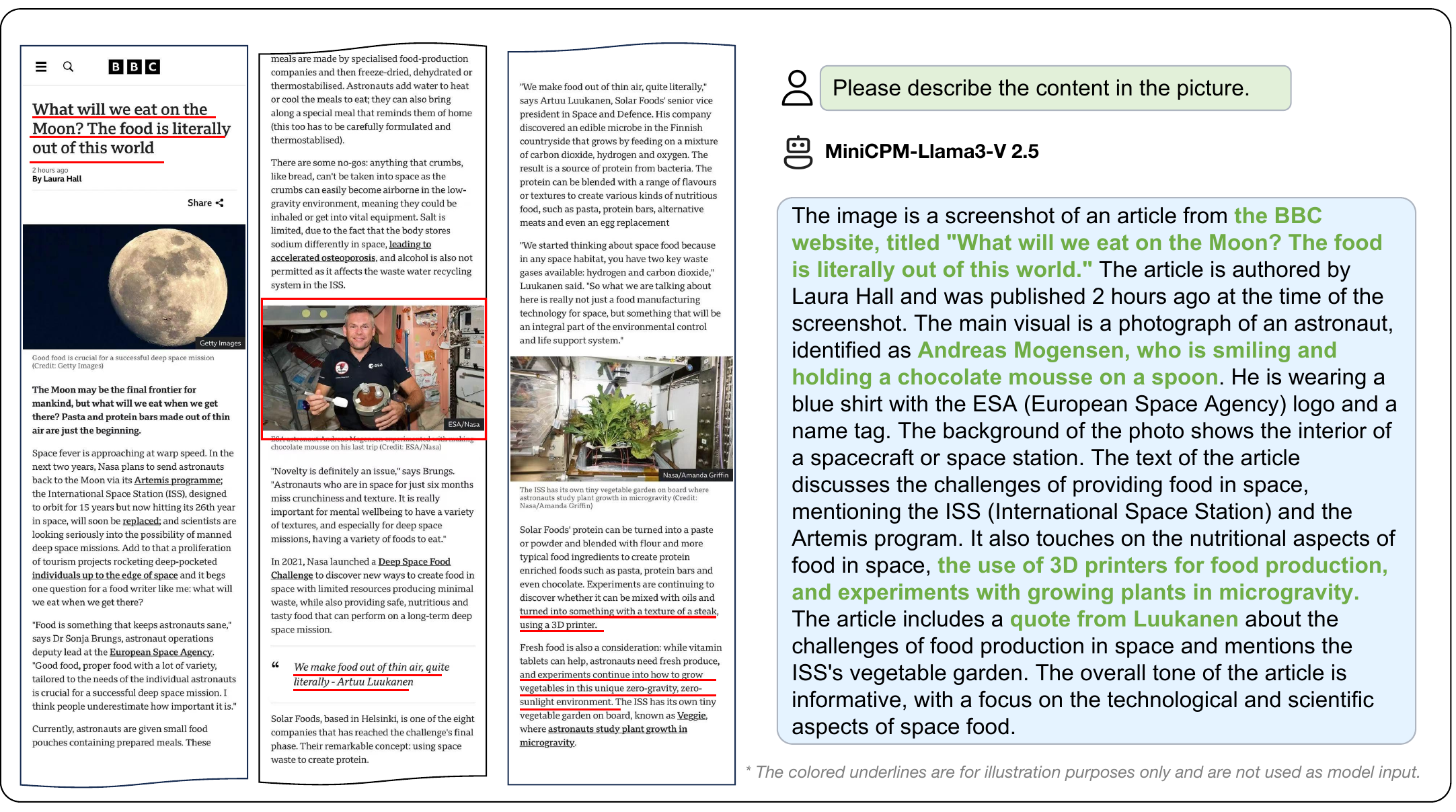}
\caption{The example of MiniCPM-Llama3-V 2.5 for the input image with extreme aspect ratio.}
\label{fig:case-long}
\end{figure*}

\begin{figure*}[!t]
\centering
\includegraphics[width=1.0\linewidth]{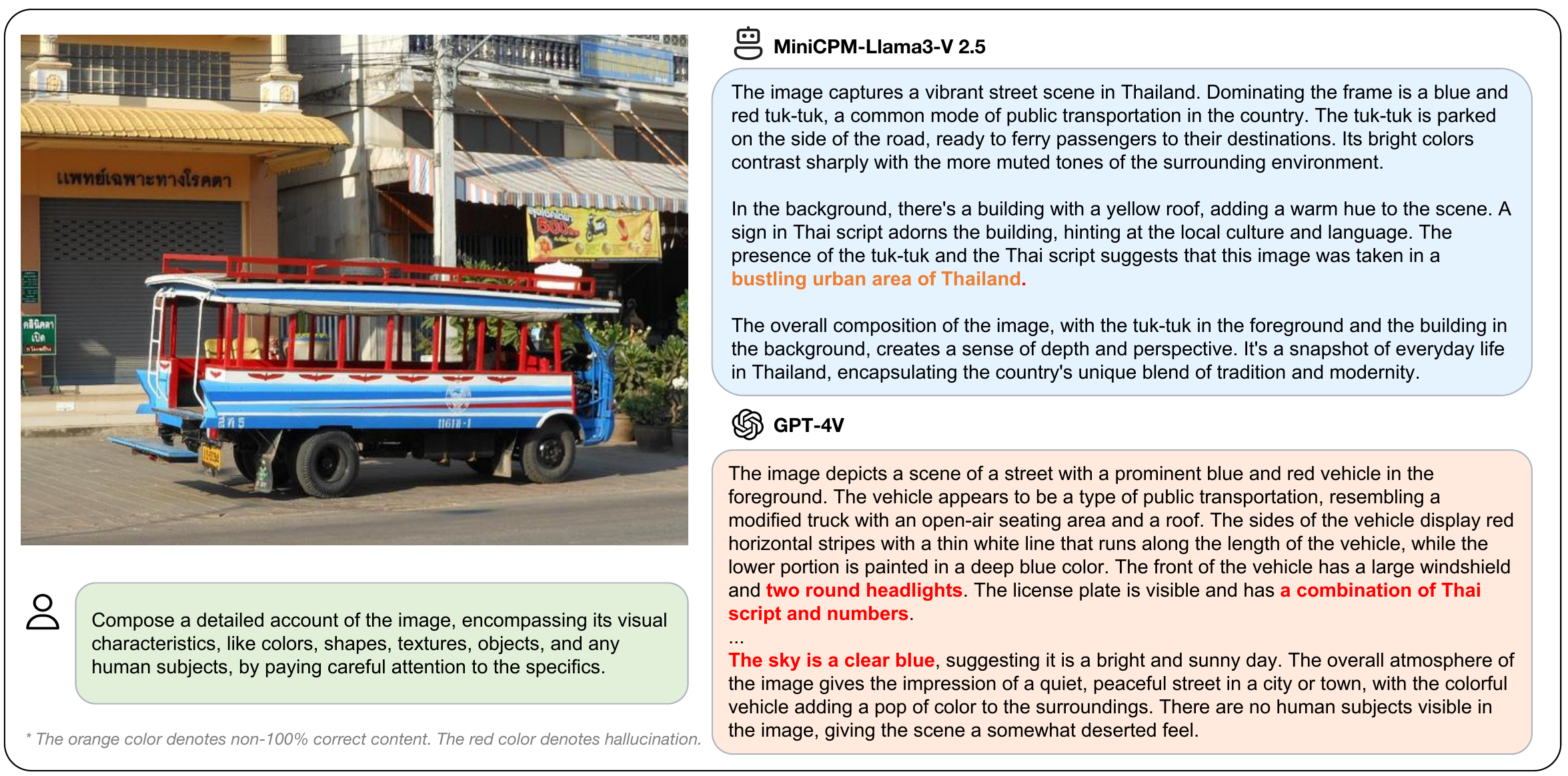}
\caption{Comparison between MiniCPM-Llama3-V 2.5 and GPT-4V on hallucinations.}
\label{fig:case-trust}
\end{figure*}

\paragraph{OCR Capability.} MiniCPM-Llama3-V 2.5 shows strong OCR capabilities in real-world scenarios. Illustrated in Fig.~\ref{fig:case-1page}, the model accurately transcribes English articles from screenshots into plain text, converts tables containing both English and Chinese content into Markdown format, comprehends code logic, and provides reasonable plans based on image content.

\paragraph{Any Aspect-ratio High-resolution Input.} A standout feature of the MiniCPM-Llama3-V 2.5 is its capability to handle high-resolution input with extreme aspect ratios. As depicted in Fig.~\ref{fig:case-long}, the model well processes input with an aspect ratio of 10:1, accurately recognizing fine-grained article contents. Interestingly, the model can also interpret images within images, correctly describing the central image as a man ``smiling and holding a chocolate mousse."

\paragraph{Multilingual Multimodal Capability.}
Benefiting from the multilingual multimodal generalization approach from VisCPM~\citep{hu2023viscpm}, MiniCPM-Llama3-V 2.5 exhibits multilingual proficiency, generalizing across more than 30 languages. Fig.~\ref{fig:multilingual-case} showcases multimodal conversations in German, French, Japanese, Korean, and Spanish, showing good knowledge of language-specific cultures.

\paragraph{Trustworthy Behavior.} Based on RLAIF-V, MiniCPM-Llama3-V 2.5 ensures more trustworthy responses with lower hallucination rates. As demonstrated in Fig.~\ref{fig:case-trust}, the model's responses exhibit fewer hallucinations as compared with powerful GPT-4V, showing its promising reliability and trustworthiness in real-world scenarios.

\section{Conclusion}

\paragraph{Contributions.} In this work, we introduce the MiniCPM-V series models as a primary exploration into powerful end-side MLLMs. Thanks to techniques such as adaptive visual encoding, multilingual generalization, and the RLAIF-V method, MiniCPM-Llama3-V 2.5 can achieve GPT-4V level performance with significantly fewer parameters. With various end-side optimization techniques, this model ensures an acceptable user experience on mobile phones.

\paragraph{Limitations.} Despite promising performance, there remain several limitations with current MiniCPM-V models. (1) Capability Depth. there is still plenty of room for improvement in enhancing multimodal understanding capability and inference efficiency. (2) Capability Width. In addition to image modality, it's promising to expand MLLM capabilities to encompass other modalities, such as video and audio, etc., where GPT-4o~\citep{openai2024gpt4o} and Google Astra~\citep{google2024astra} have given good examples.

In addition to MLLM capabilities, end-side deployment also presents unique challenges.
The inference speed and latency are still far from good enough and the model service can be limited by the battery capacity.
In addition, previous efforts on chips and deployment frameworks mainly target CNNs and LSTMs, which can be sub-optimal for MLLMs. Tailored efforts to MLLMs can bring ample room for improvement.

\paragraph{Future Works.}
Considering the current limitations and the promising future of end-side MLLMs, we anticipate increasing efforts from both academia and industry in enhancing model capabilities in terms of depth and width, and improving smartphone chips and deployment frameworks. We believe that simultaneous advancements in model capability and end-side device capacity will lead to end-side applications providing a satisfying user experience in the near future.

{
\small
\bibliographystyle{plainnat}
\bibliography{references}
}


\appendix

\begin{figure*}[t]
\centering
\includegraphics[width=1.0\linewidth]{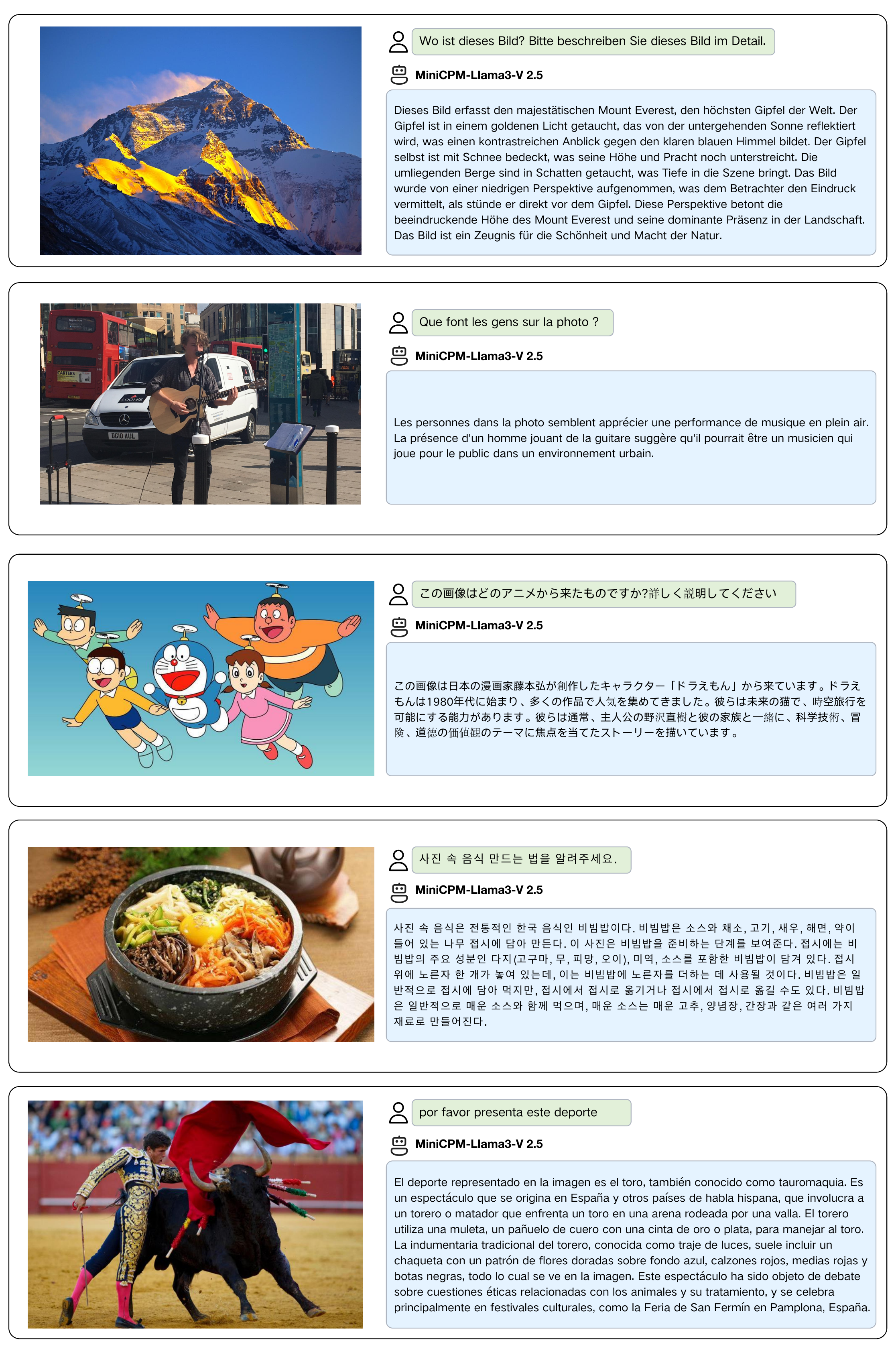}
\caption{Multilingual results of MiniCPM-Llama3-V 2.5. We select German, French, Japanese, Korean and Spanish as examples for demonstration.}
\label{fig:multilingual-case}
\end{figure*}

\begin{figure*}[t]
\centering
\includegraphics[width=1.0\linewidth]{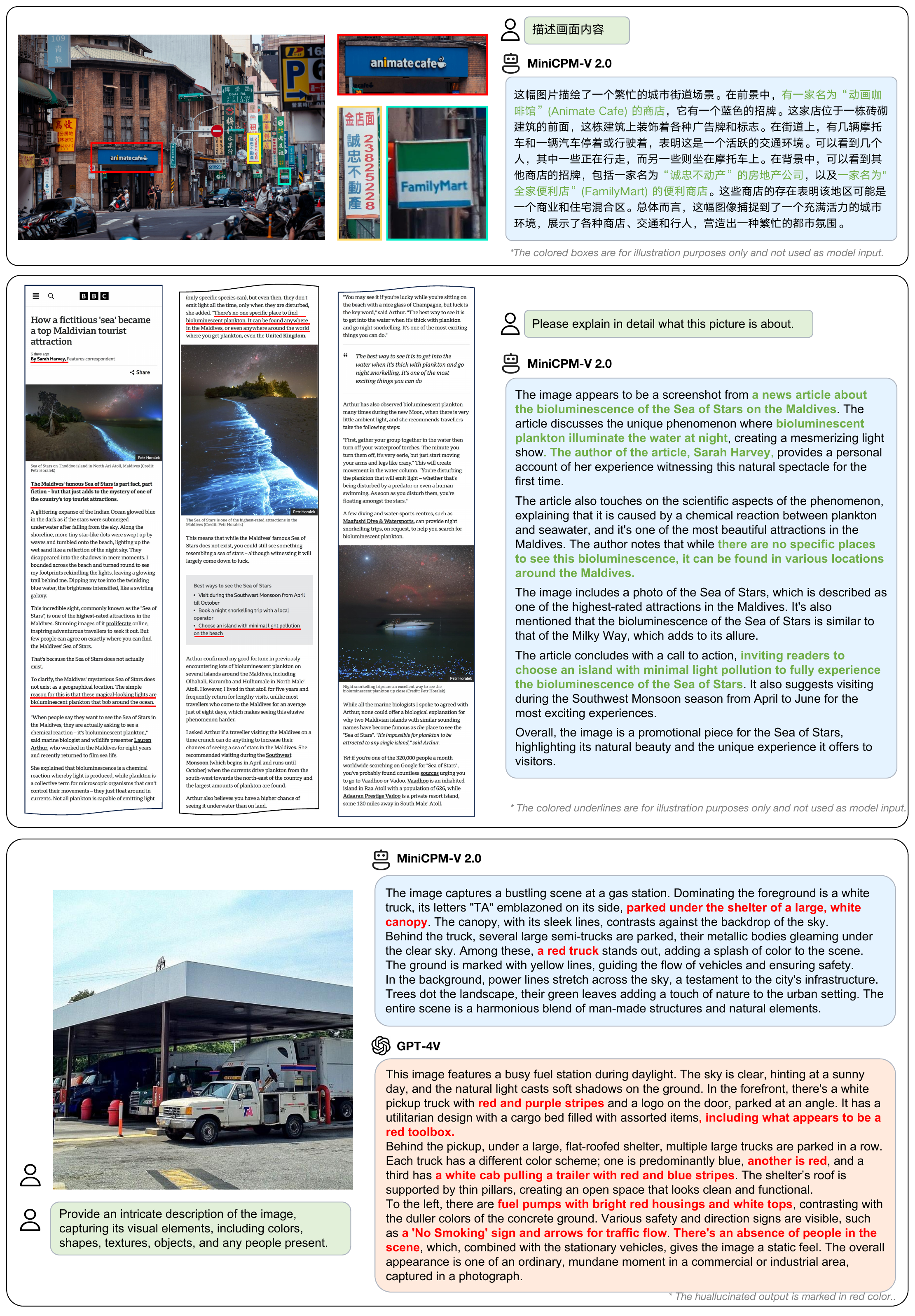}
\caption{Qualitative results of MiniCPM-V 2.0. (1) Case 1 shows a scene-text understanding example on the street. (2) Case 2 shows an example of image understanding with an extreme aspect ratio. (3) Case 3 compares MiniCPM-V 2.0 and GPT-4V on detail captioning. The hallucinated outputs are marked in red color.}
\label{fig:case}
\end{figure*}

\begin{figure*}[t]
\centering
\includegraphics[width=1.0\linewidth]{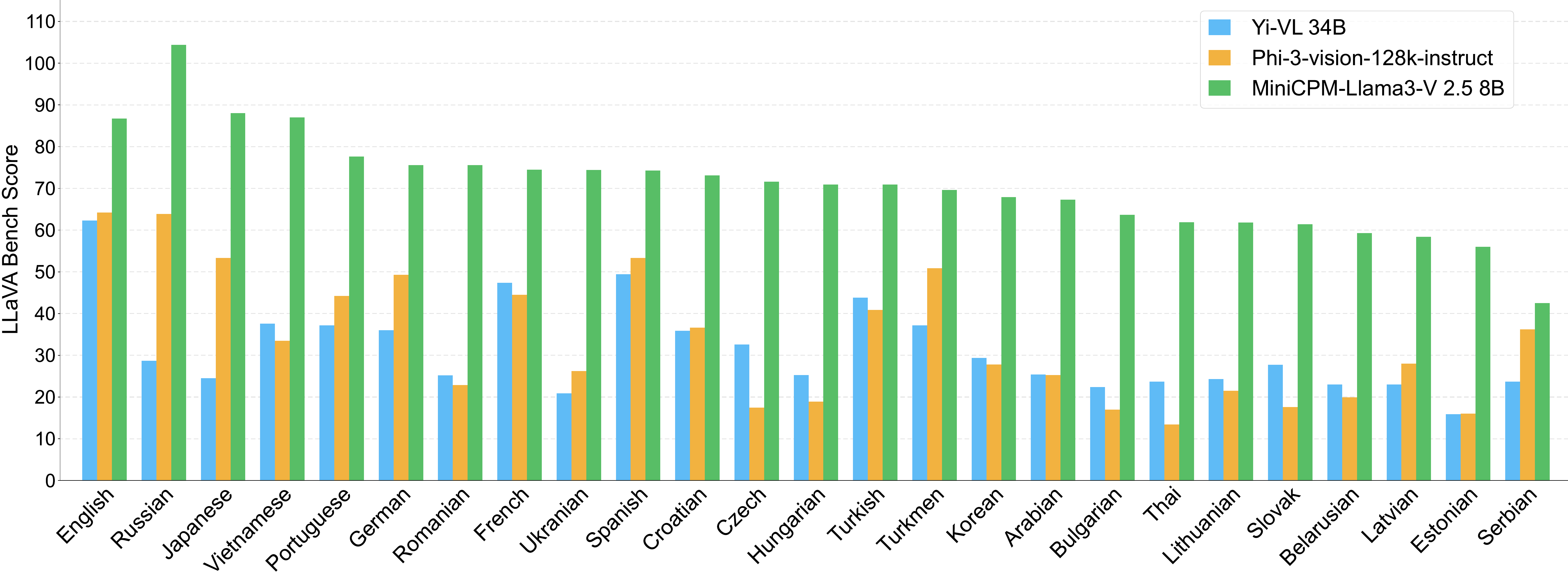}
\caption{Multilingual results. Comparison between MiniCPM-Llama3-V 2.5, Yi-VL 34B, and Phi-3-vision-128k-instruct on more languages.}
\label{fig:ml-more}
\end{figure*}

\end{document}